\documentclass[runningheads]{llncs}
\usepackage{amsmath}

\newcommand{\tcite}[1]{\scriptsize\cite{#1}}

\newcommand{\second}[1]{\textcolor{blue}{#1}}
\newcommand{\best}[1]{\textcolor{red}{#1}}

\usepackage{eccv}

\usepackage{eccvabbrv}

\usepackage{graphicx}
\usepackage{booktabs}

\usepackage[accsupp]{axessibility}  

\usepackage{hyperref}

\usepackage{orcidlink}

\begin{document}

\title{Logit Lens Supervision for Patch-Level Explanations in Vision-Language Models} 


\titlerunning{Logit Lens Supervision for Patch-Level VLM Explanations}



\author{
Parsa Esmaeilkhani\inst{1} \and
Longin Jan Latecki\inst{1}
}

\authorrunning{P. Esmaeilkhani and L. J. Latecki}

\institute{
Department of Computer and Information Sciences, Temple University,\\
Philadelphia, PA, USA\\
\email{\{parsa.esmaeilkhani, latecki\}@temple.edu}
}
\maketitle

\begin{abstract}

Modern autoregressive Vision-Language Models (VLMs) can generate fluent answers while their visual-token representations become weakly tied to the image regions from which they originate. This limits patch-level explainability: a visual token should remain interpretable as the image patch it represents. We study this issue through Logit Lens maps, obtained by projecting each visual-token embedding through the LLM vocabulary head to measure how strongly each image patch is associated with a queried textual concept, such as "cat". We introduce Logit Lens Loss (LLL), a lightweight auxiliary objective that preserves localized visual semantics by directly aligning object-related visual tokens with the vocabulary concepts describing their image regions. LLL requires no architectural modification, mask decoder, or large-scale retraining. We evaluate LLL through both explanation quality and downstream performance. Across LLaVA-v1.5-7B and Qwen2.5-VL-7B, LLL yields sharper object confidence maps, improves grounding and hallucination metrics, transfers to zero-shot pointing, and preserves VQA performance.

\keywords{Vision-Language Models \and Explainable AI \and Visual Grounding \and Logit Lens}
\end{abstract}

\section{Introduction}

\begin{figure*}[htbp]
  \centering
  \includegraphics[width=0.75\linewidth]{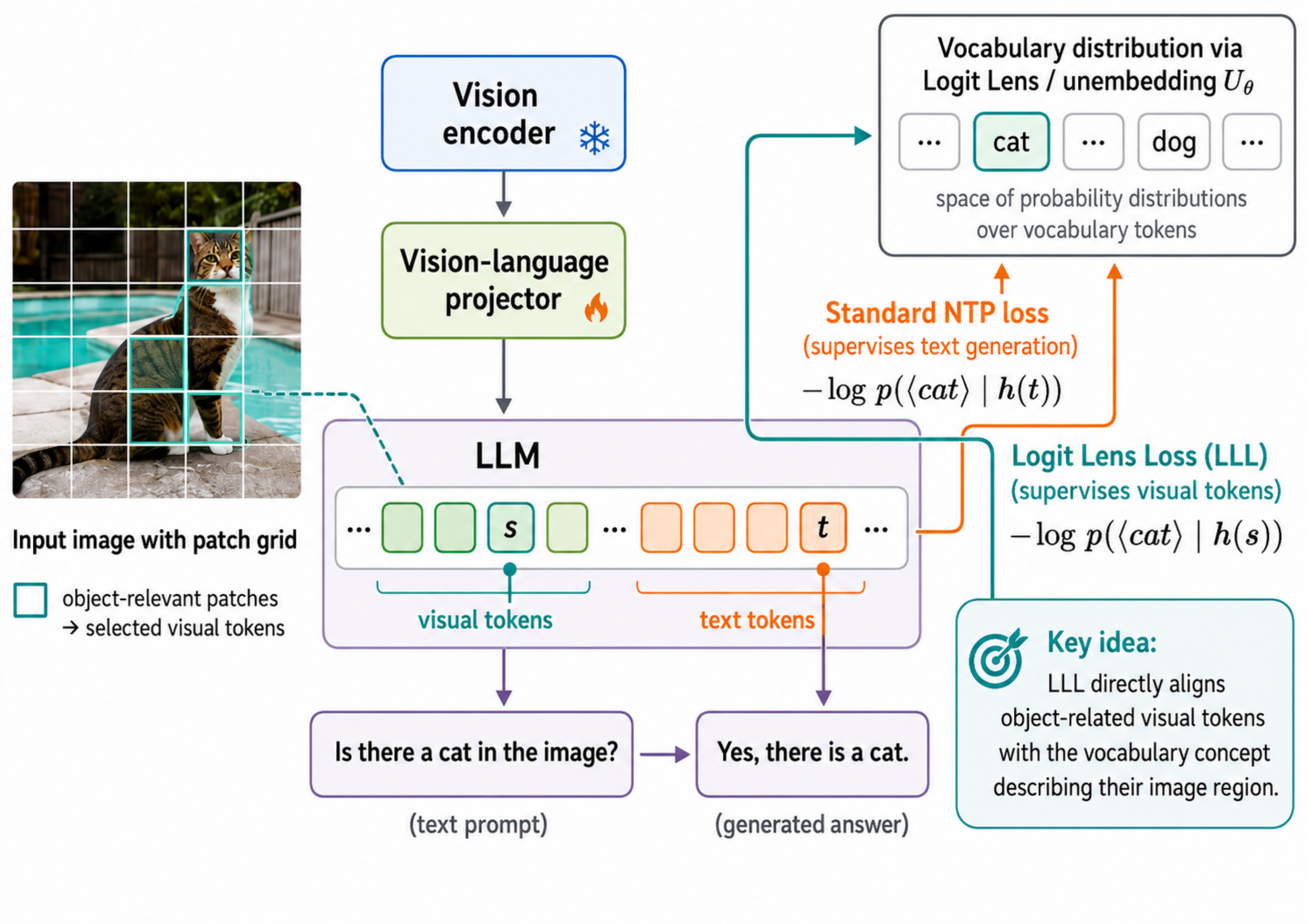}
  \caption{
  Overview of Logit Lens Loss (LLL). Standard NTP supervises text-token generation, whereas LLL directly supervises visual tokens by increasing the vocabulary probability of concepts present in their corresponding image patches. For example, a visual token s covering a cat is encouraged to predict <cat>, aligning localized patch embeddings with the relevant text concept.
  }
  \label{fig:improc}
\end{figure*}

Autoregressive Vision-Language Models (VLMs), such as LLaVA~\cite{liu2023visual},
and Qwen-VL~\cite{bai2023qwenvlversatilevisionlanguagemodel},
combine a visual encoder, typically CLIP~\cite{radford2021learning} or
DINOv2~\cite{oquab2024dinov2learningrobustvisual}, with a projection module
that maps image features into the LLM embedding space. The resulting visual
tokens are then jointly processed with the text tokens by the LLM.

As visual tokens are transformed through the layers of an LLM, their embeddings change.
In particular, the embeddings of visual tokens are mixed with the embeddings of language tokens in the attention layers of the LLM module.
However, there is a risk that information carried by visual tokens becomes diluted among the predominantly textual tokens as they propagate through the LLM layers.
In particular, the information of visual tokens may be transferred to language tokens,
as demonstrated in \cite{whatsimagedeepdCVPR2025}.
In an extreme case, the visual tokens are ignored, and the LLM can even hallucinate an image description for an empty image, as was demonstrated in  \cite{payattentionimageECCV2024,tong2024cambrian, Esmaeilkhani_2026_WACV}.
Further evidence that visual tokens are often ignored among all LLM tokens is the fact that after removing half of the visual tokens, the VLM performance does not decrease \cite{chen2024imageworth12tokens}.  In contrast, removing patches that contain the target object leads to a significant drop in performance, indicating that only a subset of visual tokens carries critical semantic information for the task \cite{neo2025interpretingvisualinformationprocessing}. 
An unintended consequence is that the connection of visual tokens to the content of corresponding image regions is lost in this process.
As pointed out in \cite{fu2025hidden},
\emph{The LLM’s ability to use its vision representations is a limiting factor in VLM performance.}

\textbf{We aim to address this issue by constraining the mixing of visual and language tokens in order to prevent the modality mixing from degrading visual information.}
This uncontrolled modality mixing often causes visual embeddings to lose their direct correspondence to the image regions from which they originated. Consequently, visual tokens may no longer encode the localized, semantic content of their source patches.
Our main contribution is \textbf{a novel visual grounding loss applied directly to visual tokens} 
to maintain semantic alignment with the language concepts describing their image regions during transformation through the LLM attention layers.
The loss constrains the embeddings of visual tokens to remain rooted in the image content as their embeddings are transformed by the LLM attention layers.

The key idea of the proposed loss is to increase the association of each visual token with the language concepts describing its content. For example, if a given image patch contains part of a cat, its visual token should yield a high probability of the vocabulary token $<cat>$. To formulate this idea, we map deep embeddings of visual tokens to probability distributions (pdf) over language vocabulary tokens using Logit Lens~\cite{logitlens2020}, see Fig.~\ref{fig:improc}. Logit Lens was originally proposed for LLMs and demonstrated that mapping a deep embedding to the pdf space is not only useful for next-token prediction, but can also reveal the content of other LLM tokens. As shown in~\cite{LogitLensICLR2025}, Logit Lens also provides insight into the representation of visual tokens in VLMs. If we focus on a selected text token, e.g., $<cat>$, Logit Lens enables us to generate object confidence maps over the input image, showing for each patch how strongly it is associated with the selected token. We propose to use these maps as a tool for visual grounding through a new loss function called \textbf{Logit Lens Loss (LLL)}. LLL increases the alignment between deep visual patch embeddings and language tokens inside a VLM by utilizing the latent language semantics of each image patch. While NTP loss increases the probability of generating a vocabulary token, e.g., $<cat>$, conditioned on previously generated text tokens, LLL increases the probability of $<cat>$ conditioned on visual tokens representing patches containing the cat. Consequently, the deep embeddings of these patches become more aligned with the deep embedding of $<cat>$.


Our approach differs from the majority of visual grounding works in that it intrinsically links visual and text tokens by making their deep embeddings more aligned, while
works like \cite{kosmos2-ICLR2023, rasheed2024glamm,lai2024lisa, wu2025flmmgroundingfrozenlarge} use the LLM predictions (or added external modules) to predict object locations as bounding box (BB) coordinates or as segmentation masks.
In these works, the LLM backbone of a VLM predicts BB coordinates without directly linking them to actual visual patches containing the target object.
So, this can be illustrated as:

\centerline{[image + text] $\ \rightarrow \ $ LLM $\ \rightarrow \ $ [text + BB as text].}

\noindent
For example, for a query "Where is a gray cat in the image?", a VLM can generate correct BB coordinates (as text) without knowing the image tokens containing the gray cat.
Of course, we can use a deterministic algorithm to draw the BB on top of the image, as it is commonly done, but the VLM is not intrinsically aware of which image patches contain the cat.
Encoding this intrinsic awareness 
into deep embeddings of visual tokens is the driving motivation of the proposed LLL.

Models finetuned with the LLL also show significant improvements in Logit Lens visualization,  reflecting improved grounding of visual token representations.
\textbf{So, these models provide better explainability than their baselines.}
Often, the Logit Lens visualization does not work well,
in particular, if one moves away from LLava, used in \cite{LogitLensICLR2025}, to newer VLMs like Qwen2.5-VL,
as demonstrated in Fig.~\ref{fig:confidence_maps}.
The reason is the diffusion of localized image information away from visual to text tokens, as we discussed above.


Our main contributions can be summarized as follows:
\begin{itemize}
    \item \textbf{Introduce a novel visual grounding loss (Logit Lens Loss, LLL):}
    A new loss is applied directly to visual tokens to preserve their connection to the underlying image content as they propagate through LLM layers. 

 \item \textbf{Embed intrinsic localization within the VLM:}
    Unlike prior work that predicts bounding boxes or masks without binding predictions to specific image patches, the proposed LLL explicitly links semantic concepts to the visual tokens that represent their true spatial locations without any architectural modifications. We visualize this with object confidence maps.

\item \textbf{Increased explainability of Models finetuned with LLL:}
    LLL improves the semantic grounding of visual token representations by providing direct supervision in the vocabulary space. This encourages visual tokens to remain aligned with the image regions they originate from, resulting in more faithfully grounded internal representations. The improvement in Logit Lens maps then follows naturally as a consequence of this enhanced grounding, as our experimental results demonstrate.

\item We show that LLL provides a direct gradient signal to visual tokens.
In contrast, the signal from NTP must propagate through the attention layers, which often results in a weak signal to the majority of visual tokens.
\end{itemize}

Overall, LLL grounds visual embeddings directly in the vocabulary space, making visual tokens more semantically aligned with the textual concepts describing their image regions while requiring no architectural modification or large-scale training. Although bounding boxes are used as lightweight supervision, LLL adds no detection head, localization module, or mask decoder. Instead, its explanations are derived directly from the VLM's own visual-token embeddings and vocabulary projection. As shown in Fig.~\ref{fig:confidence_maps}, this makes Logit Lens maps more accurate and interpretable, turning them from a weak diagnostic tool into a more faithful patch-level explanation of which visual regions the VLM internally associates with a textual concept. Importantly, LLL does not aim to teach new concepts, but to preserve localized image understanding; for example, aligning a patch with ``cat'' can also support related concepts such as ``tabby,'' ``feline,'' or ``pet,'' since these concepts are close in the LLM embedding space~\cite{mikolov2013distributedrepresentationswordsphrases}.


\vspace{-5mm}
\section{Preliminaries}

\subsection{Vision-Language Models} 
The architecture of recent autoregressive VLMs for text generation typically involves three main components: 
a vision encoder to process image inputs, 
a mapping network to project image features into image embeddings, 
and an autoregressive language model to process both image and prompt embeddings to generate text. 
We focus on two recent state-of-the-art VLMs: LLava-v1.5 ~\cite{liu2023visual} and Qwen2.5-VL ~\cite{bai2025qwen2}, 
both used in their 7B-parameter versions. 
LLava-v1.5 uses a frozen CLIP vision encoder and an MLP projection head trained on vision-language data, followed by instruction tuning. In contrast, Qwen2.5-VL employs a jointly trained vision-language architecture in which the vision encoder and projection layers are co-optimized with the language model.

The VLM input sequence typically includes visual tokens $\mathcal{P}$ from the image, followed by text tokens $\mathcal{T}$ for instructions.
These segments are concatenated to form the full input context:
$\mathbf{X}
= \big[\mathcal{P}, \mathcal{T} \big]$.
All of the visual and language tokens are passed through the LLM backbone and mixed by the attention layers in a joint embedding space, which is language-dominated \cite{wang2024mllm}, since LLMs are pretrained on a huge amount of text data.
The LLM then generates an answer sequence \(\mathcal{A}=(t_{1},\dots,t_{i-1})\) token-by-token under a left-to-right causal mask.

During generation, at each step \(i\), the model attends only to tokens at positions \(<i\) and selects the most probable next token from its vocabulary.
More precisely, an unembedding matrix $U_\theta \in \mathbb{R}^{|\mathcal{V}|\times d}$ 
maps the latent representation $h_L(t_{i-1})$ of token $t_{i-1}$
to a probability distribution over the vocabulary $\mathcal{V}$ for the prediction of next token $t_{i}$,
where 
$h_L(t_{i-1}) \in \mathbb{R}^{d}$ is obtained by transforming the
previously generated token $t_{i-1}$
through layers 1 to $L$ of the LLM backbone.
So, at each step \(i\) the model predicts \(t_i\) conditioned on the
full context \(\mathbf{X}\) and the previously generated answer tokens
\(\mathcal{A}_{a<i}=(t_{1},\dots,t_{i-1})\). 

During VLM training, the standard next-token prediction  (NTP) loss is used:
\begin{equation}
L_{\mathrm{NTP}}(\theta)
= \frac{1}{|\mathcal{A}|}\sum_{i=1}^{|\mathcal{A}|}
-\log p_{\theta}\big(t_{i}\mid \mathcal{A}_{a<i}, \mathbf{X}\big)
\end{equation}
where 
$p_{\theta}\big(t_{i}\mid \mathcal{A}_{a<i}, \mathbf{X}\big) = p_{\theta}\big(t_{i}\mid h_L(t_{i-1})\big)$ is the probability of generating the next ground truth token $t_{i} \in \mathcal{V}$ and
\(\theta\) denotes the model parameters.
We observe that the NTP loss does not include any incentives for preserving the image content initially encoded in visual tokens $\mathcal{P}$.

\subsection{Logit Lens} \label{sec:LL}

We use the Logit Lens~\cite{ LogitLensICLR2025} to reveal how visual tokens evolve into textual concepts within a VLM. Originally designed to analyze intermediate representations in language models, Logit Lens~\cite{logitlens2020} offers a direct way to project latent embeddings into the LLM’s vocabulary space.
It takes an embedding $h_l(x)  \in \mathbb{R}^{d}$ at any LLM layer and projects it through the model’s unembedding matrix 
$U_\theta: \mathbb{R}^{d} \rightarrow \mathbb{R}^{\mathcal{V}}$, 
which is the same matrix as used at the final LLM layer to predict vocabulary logits.
This gives us a logit vector over the vocabulary of text tokens: 
\begin{equation}
    U_\theta  h_l(x) = [\text{logit}_1, \text{logit}_2, \dots, \text{logit}_{|\mathcal{V}|}],
    \label{eq:logitlens}
\end{equation}
where $\text{logit}_j$ corresponds to token $j$ in the vocabulary $\mathcal{V}$. 
This vector represents the model’s predicted logit distribution over tokens after layer~$l$.

Given the logits vector obtained from the matrix-vector multiplication $U_\theta h_l(x)$, the probability distribution over the vocabulary is computed using the softmax function:
\begin{equation}
    p_\theta(v_j \mid h_l(x)) = 
    \frac{\exp(\text{logit}_j)}
         {\sum_{k=1}^{|\mathcal{V}|} \exp(\text{logit}_k)},
    \quad \forall\, v_j \in \mathcal{V}.
    \label{eq:softmax2}
\end{equation}
This yields the model’s predicted token probabilities at layer $l$, 
where $p_\theta(v_j \mid h_l(x))$ denotes the probability of token $v_j$ conditioned on the latent representation $h_l(x)$.

When applied to a visual token $x=s_i$,
the resulting probability vector $p_\theta(v_j \mid h_l(s_i))$ represents the 
distribution over vocabulary tokens that describes the visual content of 
patch $i$ as perceived by the model with parameter $\theta$ at layer $l$. 
This allows us to identify which word tokens, e.g., “cat”, “bottle”, the model associates most strongly with each patch.
So, using the Logit Lens on visual tokens directly relates the corresponding image patches to textual concepts. 

\section{Methodology: Logit Lens Loss}

As our confidence maps demonstrate in Fig. ~\ref{fig:confidence_maps}, VLMs only partially focus on the correct image patches corresponding to an object category. This occurs because VLMs finetuned with next-token prediction receive only weak supervision regarding the visual patterns present in the image. This observation motivates us to explore how to teach VLMs to ensure that visual tokens remain semantically tied to the image regions they represent as they propagate through the language model layers, without sacrificing the language modeling capabilities of the LLM component. 

While NTP optimizes the model for textual coherence, it does not directly constrain the internal representations of visual tokens, allowing them 
to drift away from their image-related content 
as attention layers mix visual and textual embeddings. To address this, we introduce a grounding objective that explicitly links each visual token to the textual concepts describing its underlying image content. 

Our approach builds on the idea that every deep embedding $h_l(x)$ within the LLM implicitly encodes a distribution over vocabulary tokens via the Logit Lens projection $U_\theta h_l(x)$. By treating this distribution as the “semantic fingerprint’’ of a patch, we define a loss that encourages visual tokens corresponding to object regions to produce vocabulary distributions consistent with their ground-truth labels (e.g., high probability for the word “cat’’ inside a cat region, and low elsewhere). Meanwhile, the loss discourages patches unrelated to a given object category from developing unrelated connections with that category. In essence, it drives each patch to focus more on objects present within its region and less on those absent. 
This yields a simple yet effective auxiliary objective, termed Logit Lens Loss, that complements the standard NTP loss by grounding visual embeddings directly in the vocabulary space.

Let $\mathcal{P}$ denote the set of all visual (patch) tokens in the image and 
$\mathcal{P'} \subseteq \mathcal{P}$ the subset of patches inside the ground-truth region (e.g.,
bounding box) of the queried object (when present). In other words, we obtain $\mathcal{P'}$ by projecting the ground-truth bounding box onto
the visual-token grid and selecting tokens whose receptive fields overlap with
the annotated object region.


Following common practice in visual grounding tasks, we use bounding box annotations to indicate which visual tokens correspond to the target object during training \cite{xiao2024towards}. 
Similar supervision is used in recent VLMs for grounding, which predict bounding box coordinates as text, employ segmentation heads for mask prediction, or use point annotations for object localization \cite{kosmos2-ICLR2023,rasheed2024glamm,deitke2025molmo}. 
In contrast, we use bounding boxes only to supervise visual token representations, without introducing explicit localization outputs or architectural modifications.

For a target vocabulary token $v$ and layer-$l$ representation $h_l(s)$, 
\textbf{Logit Lens Loss (LLL)} is defined as
\begin{align}
\label{eq:LLL}
L_{\mathrm{LLL}}(\theta) &= 
\frac{1}{|\mathcal{P'}|} \sum_{s \in \mathcal{P'}}
-\log p_{\theta}(v \mid h_l(s)) \\
&+ \frac{1}{|\mathcal{P}\setminus \mathcal{P'}|} \sum_{s \in \mathcal{P} \setminus \mathcal{P'}}
-\log \big(1 - p_{\theta}(v \mid h_l(s))\big). \nonumber
\end{align}


When the queried concept spans multiple tokens (e.g., subword splits or multi-word phrases), we use a token set $V \subseteq \mathcal{V}$ and apply the loss independently to each token, averaging the results. 
For example, in “a woman in a blue shirt,” we apply the loss to key tokens such as “woman,” “blue,” and “shirt,” selected via simple part-of-speech filtering (e.g., nouns and adjectives).

For simplicity, let's say $v=\{\mathrm{<cat>}\}$ is a vocabulary token representing the word "cat" and 
$\mathcal{P'}$
is a subset of visual tokens corresponding to image patches depicting the cat in the input image, 
then Eq.~\eqref{eq:LLL} reduces to
\begin{equation}
L_{\mathrm{LLL}}(\theta)=\frac{1}{|\mathcal{P'}|}\sum_{s \in \mathcal{P'}}
-\log p_{\theta}\bigl(\mathrm{<cat>} \mid h_l(s)\bigr).
\end{equation}
It aims at increasing the probability of token $\mathrm{<cat>}$ in the vocabulary distribution for each deep embedding $h_l(s)$ of image patch $s \in \mathcal{P'}$ that depicts the cat in the image.
We stress that this does not suppress other concepts like "kitten" or "cat ear".
The expressive power of a VLM comes from its pre-training; the LLL only helps to preserve the image-grounded locality of visual deep embeddings, and does not aim at teaching new concepts to VLMs.

The second term in Eq.~\eqref{eq:LLL}
aims at decreasing the probability of token $\mathrm{<cat>}$ in the vocabulary distribution for each deep embedding $h_l(s)$ of image patch $s \in \mathcal{P} \setminus \mathcal{P'}$ outside the cat bounding box.



We propose to use the $L_{\mathrm{LLL}}$ as the auxiliary loss in addition to the next token prediction loss $L_{\mathrm{NTP}}(\theta)$ of the VLM
to optimize the VLM's parameters for both objectives:
\begin{equation} \label{eq:totalloss}
L_{\text{total}} = L_{\text{NTP}} + \lambda L_{\text{LLL}}.  
\end{equation}
We empirically set $\lambda$ to 0.5, which we found effective across all tasks. 
Like $L_{\text{NTP}}$, the proposed $L_{\text{LLL}}$ is only applied to the last VLM layer, i.e., $l=L$ in Eq.~\eqref{eq:LLL}.
Moreover, after finetuning with LLL,
it is also sufficient to apply Logit Lens to the last layer.
In contrast, in \cite{ LogitLensICLR2025},
Logit Lens is computed as the maximum over all layers.

Both terms in Eq.~\eqref{eq:totalloss} are nicely aligned since both are based on increasing the probabilities of generating selected vocabulary tokens, as illustrated in Fig.~\ref{fig:improc}.
Formally, since both visual and textual embeddings can be projected into probability distributions over the vocabulary via the Logit Lens, they share a common semantic space. We denote this shared space with $\mathcal{F}$ of all probability distribution functions (pdfs)
$p_\theta(\cdot \mid h_l(x)): \mathcal{V} \rightarrow [0,1]$
over the vocabulary tokens $\mathcal{V}$,
i.e., 
\begin{equation}
\mathcal{F} = \{ p_\theta(\cdot \mid h_l(x)) \ \ | \ x \in \mathbb{R}^d \}. 
\end{equation}
Let $f:\mathbb{R}^d \rightarrow \mathcal{F}$ 
be defined as $f = \mathrm{softmax} \circ U_\theta$
following the process in Sec.~\ref{sec:LL}.
Space $\mathcal{F}$ is a common space to link the visual and language tokens, since 
$f(s) \in \mathcal{F}$ and $f(t) \in \mathcal{F}$ for all visual tokens $s\in \mathcal{P}$ and text tokens $t\in \mathcal{T}$.

For example, if a given visual token $s$ corresponds to an image patch containing part of a cat, we want 
$p_\theta(<cat> \mid h_l(s))$ to have a high value, which is analogous to increasing the probability of generating token $t_{i} = <cat>$ as the next ground truth token, which is given by
$p_{\theta}\big(t_{i}\mid \mathcal{A}_{a<i}, \mathbf{X}\big) = p_{\theta}\big(t_{i}\mid h_L(t_{i-1})\big)$
in the next-token prediction loss $L_{\text{NTP}}$.

LLL applies the loss directly to visual tokens, while
NTP applies loss directly only at text positions,
Minimizing $L_{\text{LLL}}$
applies cross-entropy through the unembedding matrix to visual tokens, whose gradient explicitly pulls the patch embedding $h_L(s)$ 
toward the unembedding vector of the target word $v$, thereby enforcing semantic alignment in the same space the LLM uses to generate language.
We focus on finetuning to demonstrate the effectiveness of LLL without requiring large-scale retraining, although the proposed loss is general and could also be applied during pre-training.

\section{Gradient Strength}
We show that 
\textbf{the gradient of $L_{LLL}$ provides a direct signal to visual tokens}.
For simplicity, let us consider a single positive visual token $\mathcal{P'} = \{s\}$ in the last layer $ L$. Then $L_{LLL}$ reduces to \begin{equation}
L_{\mathrm{LLL}}(s)
= - \log p_\theta\!\left(v \mid h_L(s)\right).
\end{equation}
Let
$z = U h_L(s)$, 
where $U$ is the unembedding matrix,
$p = \mathrm{softmax}(z)$,
and $p_v = p_\theta(v\mid h_L(s))$.
Using the standard softmax cross-entropy derivative,
\begin{equation}
\frac{\partial}{\partial z}\big(-\log p_v\big) = p - e_v,
\end{equation}
where $e_v$ is the one-hot vector for token $v$ and vector $p\in\mathbb{R}^{|V|}$ denotes the softmax probability distribution
over the vocabulary:
$p = \mathrm{softmax}\!\big(U h_L(s)\big)$,
$\sum_{j=1}^{|V|} p_j = 1$.
By the chain rule,
\begin{equation}
\frac{\partial L_{\mathrm{LLL}}(s)}{\partial h_L(s)}
= U^\top (p - e_v)
=\sum_{j} p_j\,U_j^\top - U_v^\top,
\end{equation}
where scalar $p_j$ is the $j$-th component of $p$, corresponding to the probability
assigned to vocabulary token $j$.

We obtain that minimizing $L_{\mathrm{LLL}}$ directly pushes the
last-layer patch representation $h_L(s)$ toward the unembedding direction
corresponding to the target vocabulary token $v$.

\textbf{In contrast, NTP provides only a weak signal to the majority of visual tokens.}  We provide the detailed derivation in the appendix.

{
\captionsetup[subfigure]{labelformat=empty,font=footnotesize}

\newlength{\confmapheight}
\setlength{\confmapheight}{20mm}

\newcommand{\confimg}[1]{%
  \includegraphics[width=\linewidth,height=\confmapheight,keepaspectratio]{#1}%
}

\newcommand{\confcellwithtext}[2]{%
  \begin{minipage}[t]{\linewidth}
    \vspace{0pt}
    \centering
    \confimg{#1}\\[-0.3mm]
    {\tiny #2}
  \end{minipage}%
}

\newcommand{\confcell}[1]{%
  \begin{minipage}[t]{\linewidth}
    \vspace{0pt}
    \centering
    \confimg{#1}
  \end{minipage}%
}

\begin{figure*}[!t]
  \centering
  \setlength{\tabcolsep}{0pt}
  \renewcommand{\arraystretch}{1.0}

  \begin{tabular}{@{}
    p{0.235\textwidth}
    @{\hspace{0.01\textwidth}}
    p{0.235\textwidth}
    @{\hspace{0.01\textwidth}}
    p{0.235\textwidth}
    @{\hspace{0.01\textwidth}}
    p{0.235\textwidth}
    @{}}

    \confcellwithtext{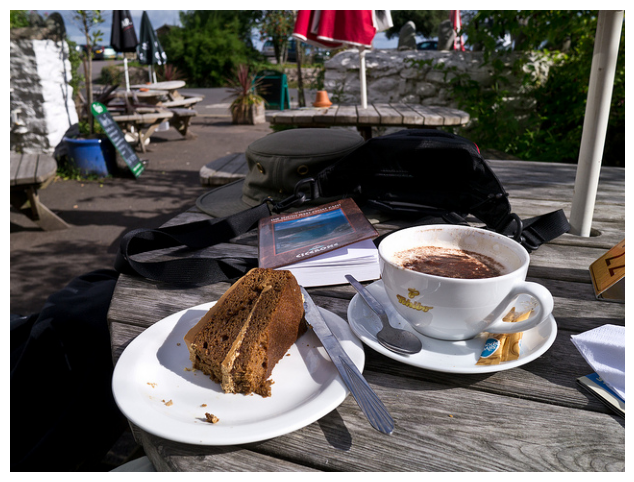}
      {``Slice of \textcolor{red}{cake} next to cup''}
    &
    \confcell{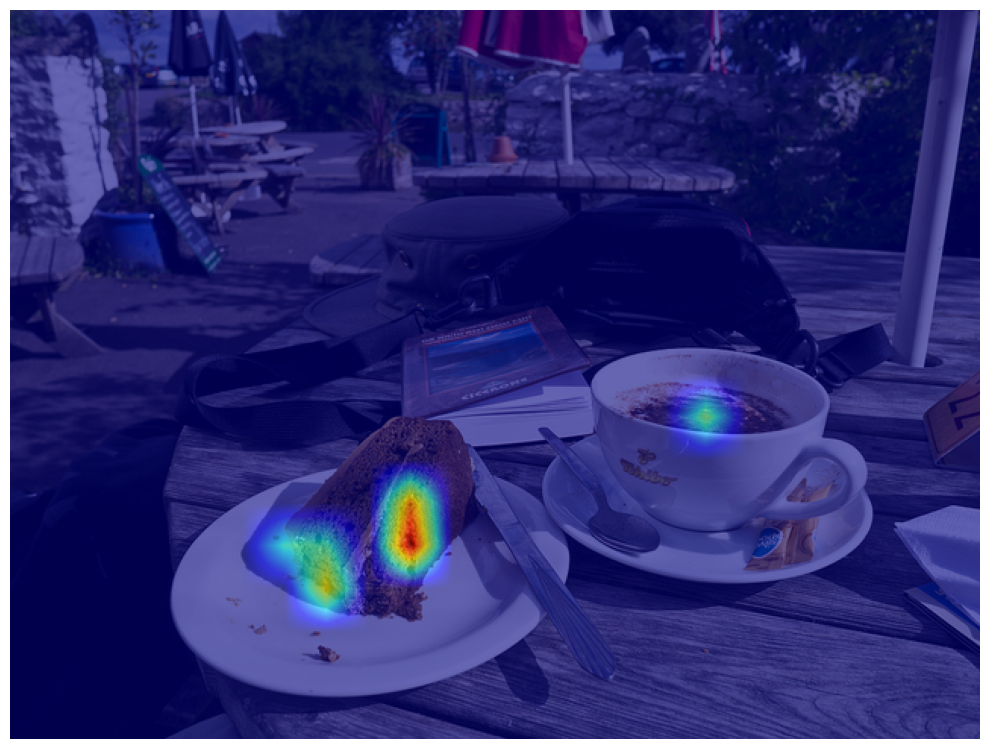}
    &
    \confcell{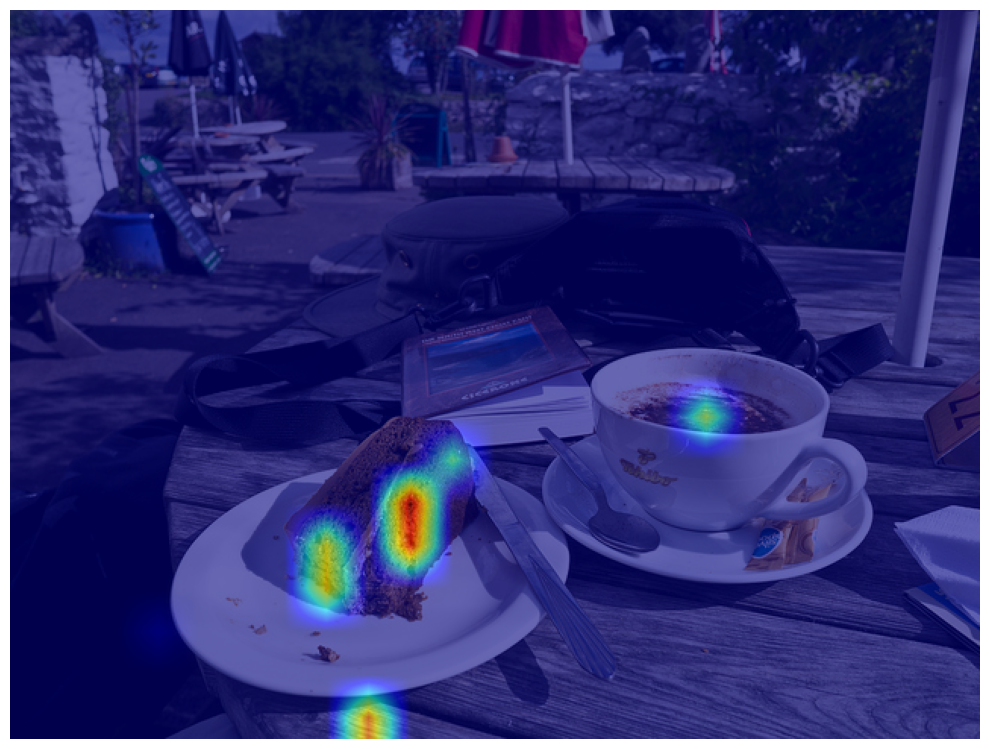}
    &
    \confcell{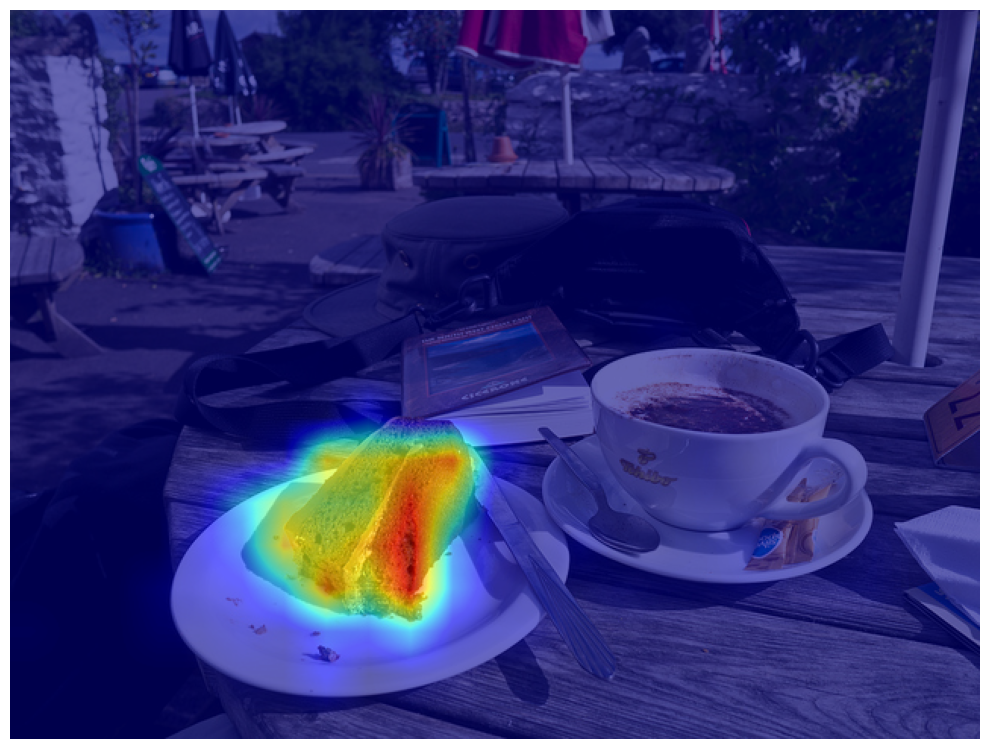}
    \\[1.5mm]

    \confcellwithtext{figures/confidence_maps/cake.png}
      {``\textcolor{red}{Book} behind cake''}
    &
    \confcell{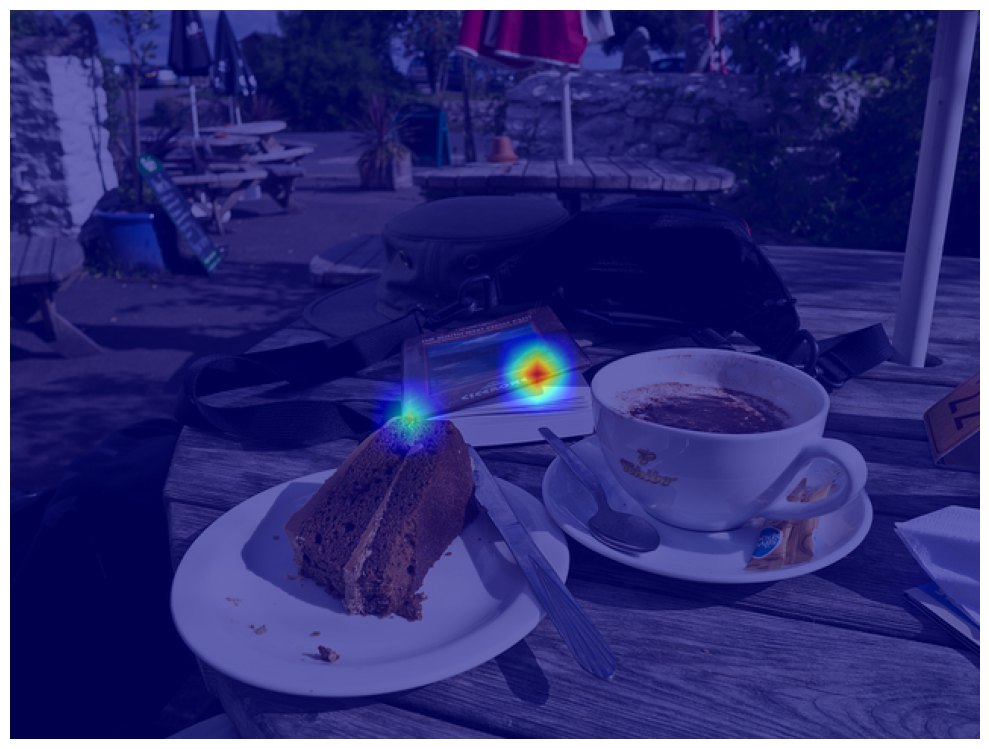}
    &
    \confcell{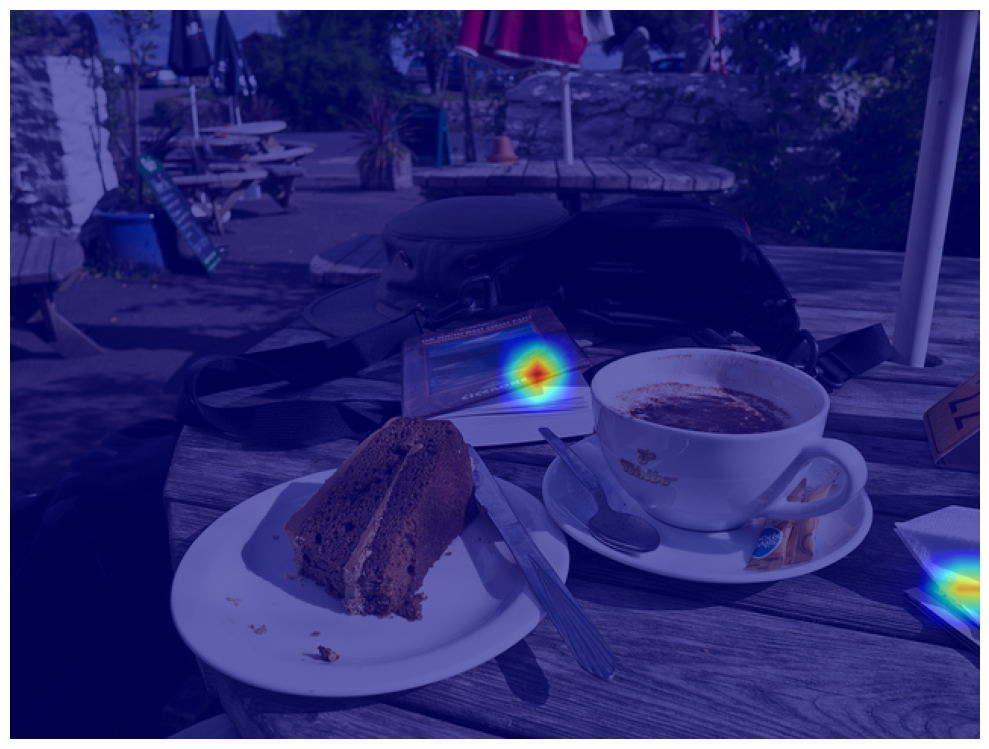}
    &
    \confcell{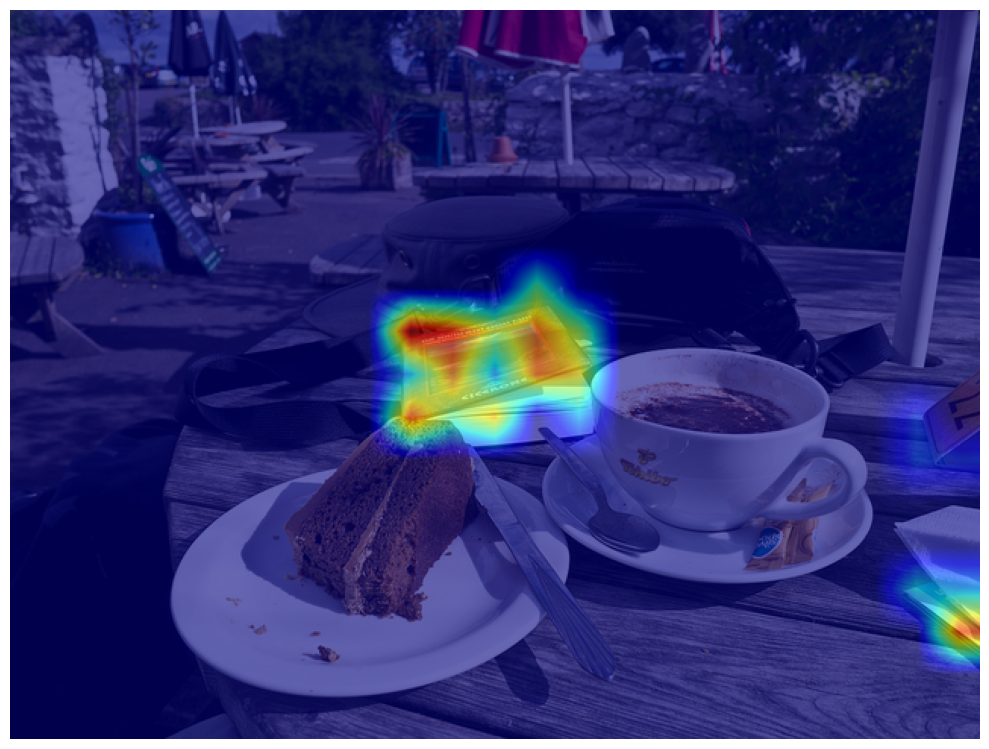}
    \\[1.5mm]

    \confcellwithtext{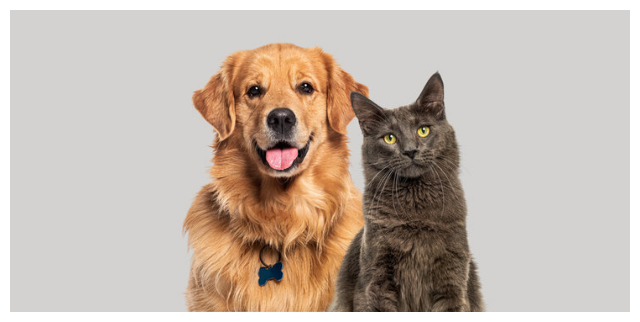}
      {``\textcolor{red}{Dog} next to cat''}
    &
    \confcell{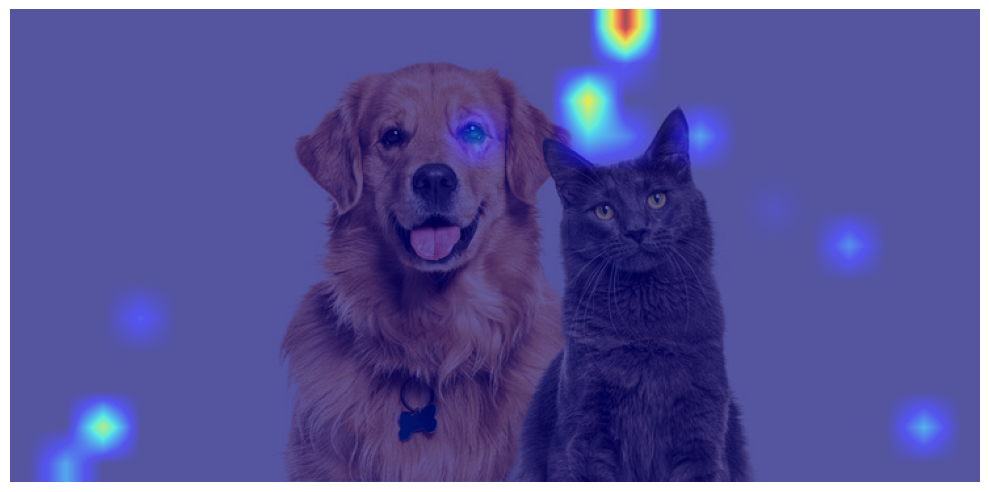}
    &
    \confcell{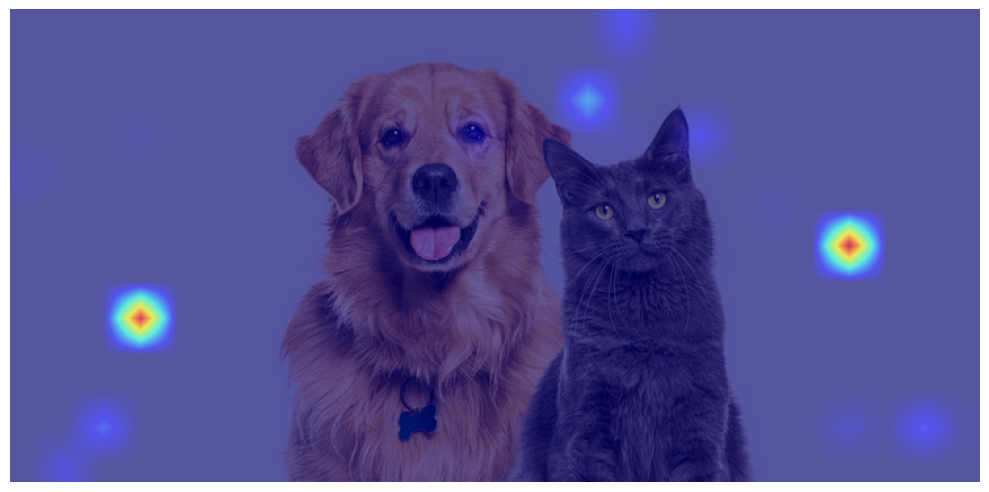}
    &
    \confcell{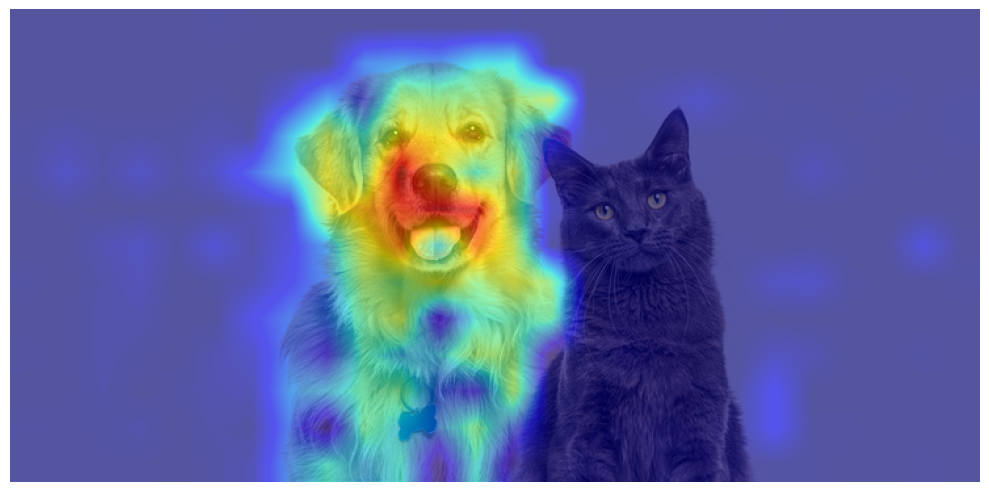}
    \\[1.5mm]

    \confcellwithtext{figures/confidence_maps/cat_dog.png}
      {``\textcolor{red}{Cat} beside dog''}
    &
    \confcell{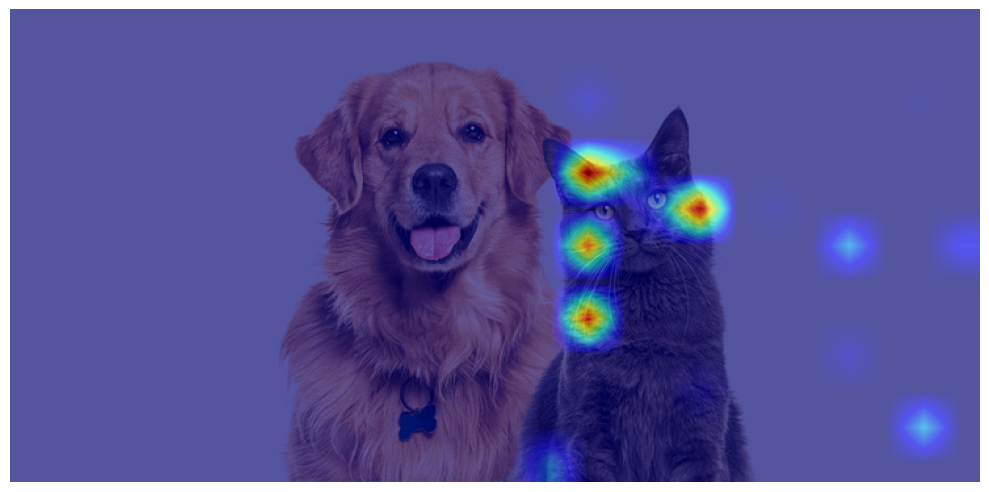}
    &
    \confcell{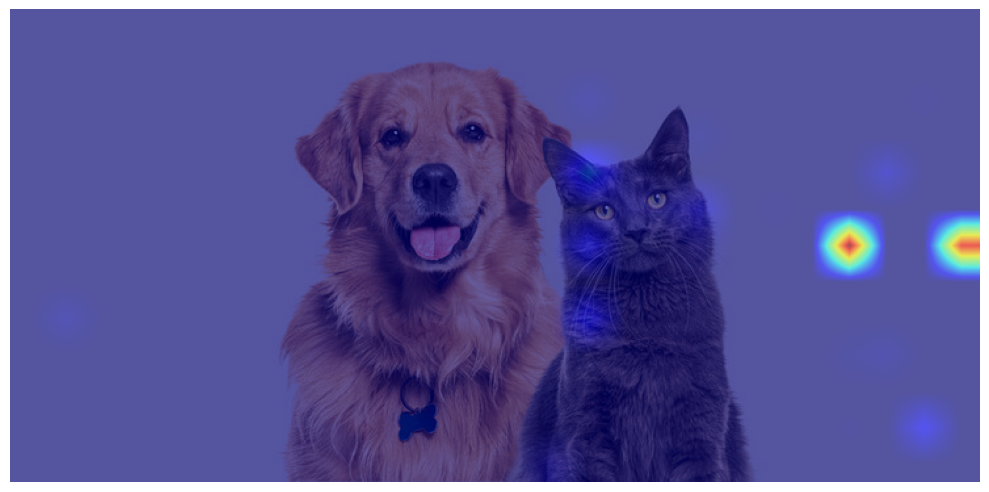}
    &
    \confcell{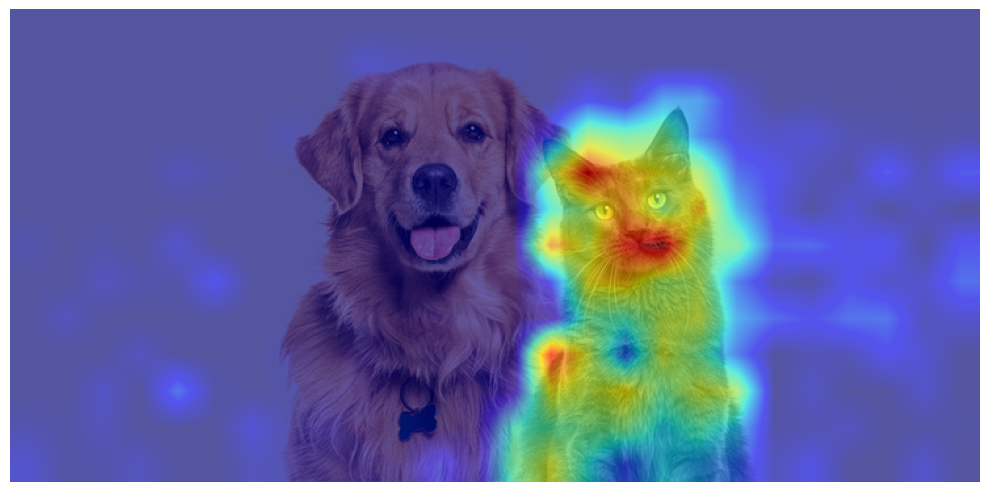}
    \\[1.5mm]

    \confcellwithtext{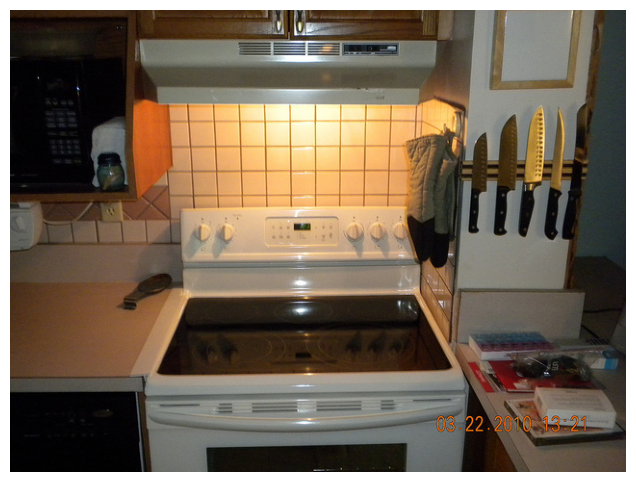}
      {``\textcolor{red}{Knife} in  holder''}
    &
    \confcell{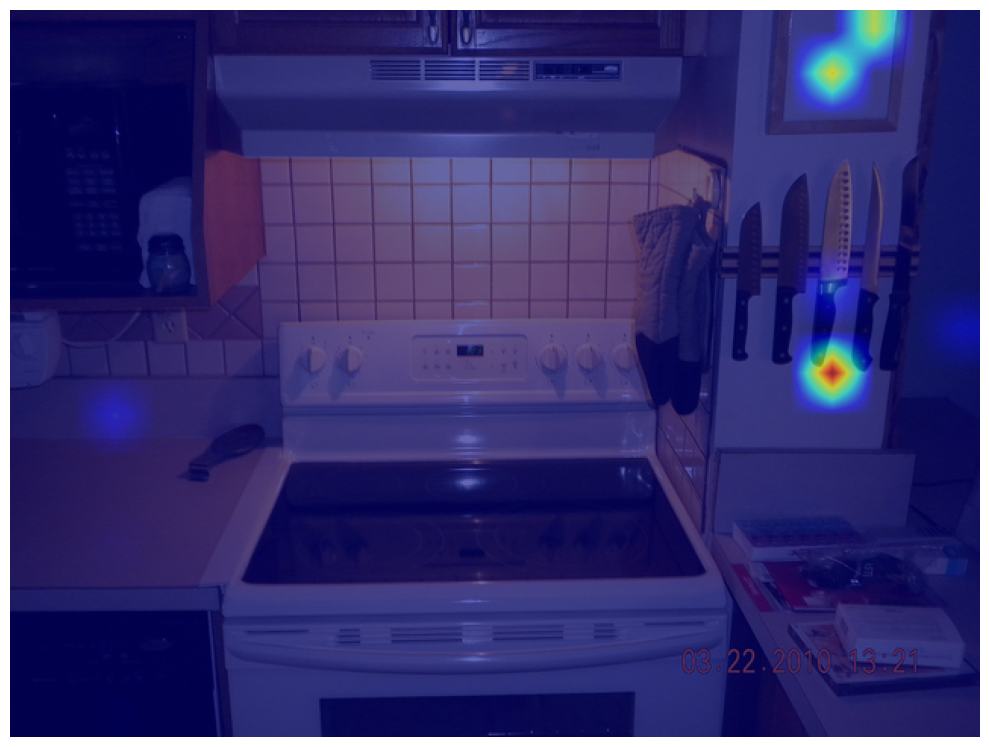}
    &
    \confcell{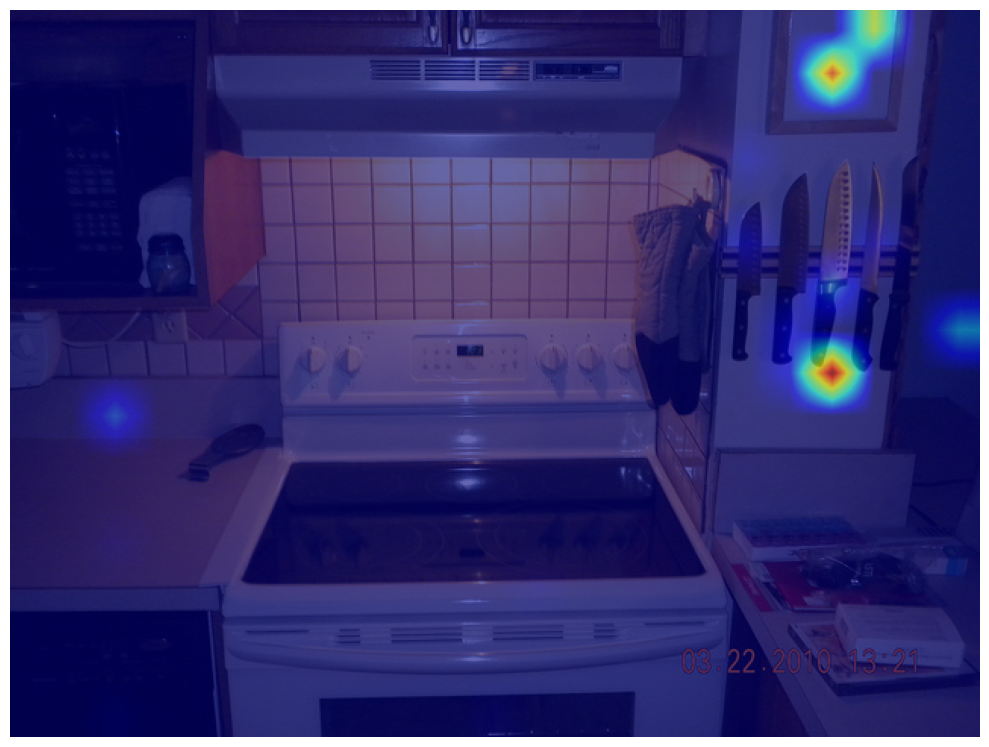}
    &
    \confcell{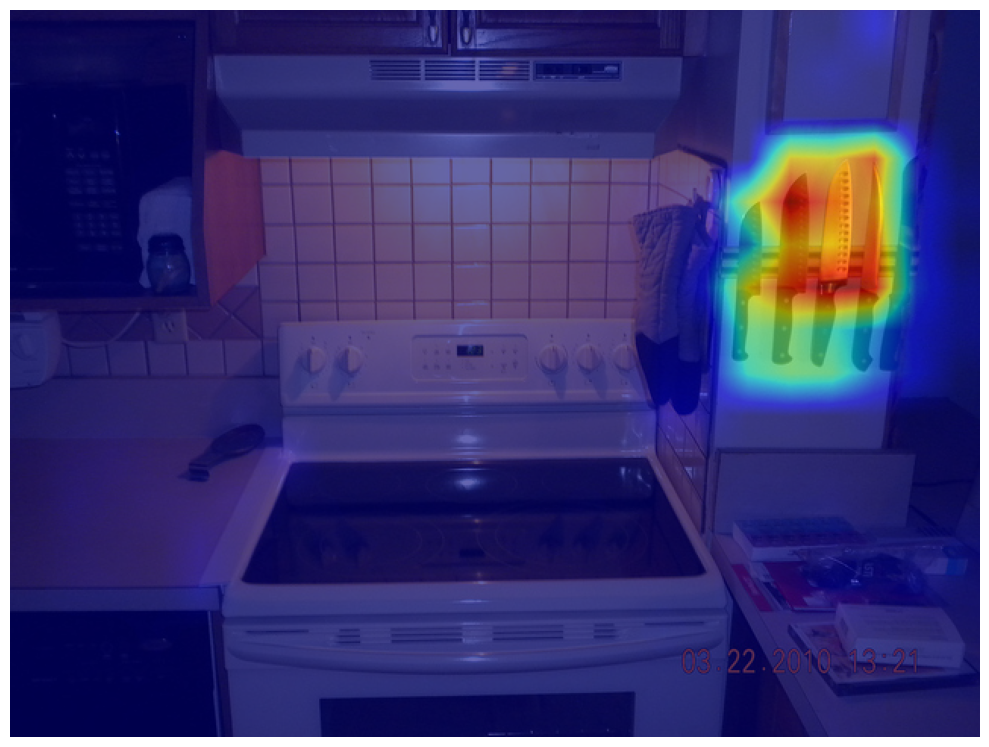}

  \end{tabular}

  \caption{Logit Lens object confidence maps produced by Qwen2.5-VL-7B.
First column: input image with the referring expression, where the queried object is highlighted in red.; second: base model; third: NTP-finetuned; fourth: NTP+LLL-finetuned.
Compared to base and NTP models, the NTP+LLL model preserves patch-level semantics despite cross-modal mixing in the LLM, yielding sharply localized, object-aligned Logit Lens maps.}
  \label{fig:confidence_maps}
  \vspace{-5mm}
\end{figure*}
}

\vspace{-3mm}
\section{Experimental Evaluation}
\vspace{-3mm}

The main benefit of finetuning with LLL is a huge increase in the quality of Logit Lens object confidence maps, reflecting improved semantic alignment of visual tokens with their corresponding image regions, as demonstrated visually in Fig.~\ref{fig:confidence_maps}.
The second column shows the maps of the base model, the third NTP, and the fourth NTP+LLL finetuned.
LLL changes sparse and confusing signals into sharper, precisely localized object confidence maps, demonstrating its beneficial impact on Logit Lens visualization of referred objects. We further validate the learned grounding through RES, POPE, zero-shot PixMo-Points, and VQA.


In the first two rows, although the scene within the image is cluttered, the model still successfully localizes the book and the cake, despite many distracting background details.
Similarly, in rows three and four, when the object text token changes from dog to cat, the deep embeddings clearly reflect this shift: the patches corresponding to the cat appear brighter, indicating that the model effectively distinguishes between the two objects. We further evaluate LLaVA-v1.5-7B and Qwen2.5-VL-7B-instruct base and finetuned models on visual grounding, object hallucination, and VQA tasks.

\begin{table}[t]
\centering
\scriptsize
\setlength{\tabcolsep}{2pt}
\renewcommand{\arraystretch}{0.92}
\caption{RES results of supervised and training-free VLM-based methods in cIoU (\%). The first two blocks evaluate masks from Logit Lens maps; SAM denotes frozen SAM postprocessing. Red/blue denote best/second-best.}
\vspace{-2mm}
\begin{tabular}{@{}lccc@{}}
\toprule
\textbf{Method} & \textbf{RefCOCO} & \textbf{RefCOCO+} & \textbf{RefCOCOg} \\
\midrule

\multicolumn{4}{@{}l}{\textit{LLaVA-v1.5-7B}}\\[0.2mm]
\hspace{0.8em} Base~{\tiny\cite{LogitLensICLR2025}} & 59.4 & 48.7 & 49.5 \\
\hspace{0.8em} NTP & 61.2 & 50.3 & 51.1 \\
\hspace{0.8em} NTP+LLL & 67.8 & 58.6 & 60.2 \\
\hspace{0.8em} NTP+LLL+SAM & 75.4 & 66.9 & 68.7 \\

\midrule
\multicolumn{4}{@{}l}{\textit{Qwen2.5-VL-7B-Instr.}}\\[0.2mm]
\hspace{0.8em} Base~{\tiny\cite{LogitLensICLR2025}} & 63.2 & 52.4 & 53.2 \\
\hspace{0.8em} NTP & 65.1 & 53.7 & 54.5 \\
\hspace{0.8em} NTP+LLL & 71.2 & 63.1 & 65.9 \\
\hspace{0.8em} NTP+LLL+SAM & \second{80.1} & 72.6 & \second{74.3} \\

\midrule
\multicolumn{4}{@{}l}{\textit{Finetuned VLMs}}\\[0.2mm]
LISA-7B~{\tiny\cite{lai2024lisa}} & 74.1 & 62.4 & 66.4 \\
GSVA-7B~{\tiny\cite{xia2024gsvageneralizedsegmentationmultimodal}} & 76.4 & 64.5 & 71.1 \\
F-LMM-7B~{\tiny\cite{wu2025f}} & 75.2 & 63.7 & 67.1 \\
LISA-13B~{\tiny\cite{lai2024lisa}} & 73.4 & 62.3 & 68.2 \\
GSVA-13B~{\tiny\cite{xia2024gsvageneralizedsegmentationmultimodal}} & 77.7 & 68.0 & 73.2 \\
GLaMM~{\tiny\cite{rasheed2024glamm}} & 79.5 & \best{75.9} & \best{76.8} \\
PSALM~{\tiny\cite{zhang2024psalm}} & \best{83.6} & \second{72.9} & 73.8 \\
\bottomrule
\end{tabular}
\label{tab:res}
\vspace{-2mm}
\end{table}
\subsection{Referring Expression Segmentation} \label{sec:RES}
In order to quantify the quality of generated Logit Lens object confidence maps, we consider the task of referring expression segmentation (RES) in Table~\ref{tab:res}. We obtain segmentation masks either by thresholding the confidence map, or by extracting a bounding box from the map and using it to prompt a frozen segmentation model to generate a mask.
The first three rows for LLaVA-v1.5 and Qwen2.5-VL show segmentation results obtained by thresholding the Logit Lens heat maps. We observe a large improvement when models are finetuned with NTP + LLL compared to NTP alone (second vs.\ third row). Moreover, when the results of the third row are postprocessed with frozen SAM~\cite{SAM2023} (fourth row), performance becomes comparable to SOTA. In particular, \textbf{Qwen2.5-VL-7B finetuned with NTP+LLL achieves second-best performance on RefCOCO and RefCOCOg}, and remains close to SOTA on RefCOCO+ after postprocessing with SAM.

We train on a mixture of datasets combining the RefCOCO/+/g train splits~\cite{yu2016modeling} with LLaVA-Instruct-150K \cite{liu2023visual}. The RefCOCO series, built on MS COCO images, requires models to localize objects referred to by natural language descriptions. This mixture supports both grounding capability and preservation of language ability. We report results on the validation sets following the standard protocol.

\vspace{-6mm}
\subsection{POPE: Polling-based Object Probing} 
\vspace{-2mm}

POPE~\cite{POPE2023} was originally designed to make the VQA evaluation fair and simple.
It evaluates VLMs using binary “yes/no” object queries instead of caption generation. Each question follows the schema “Is there a $<$object$>$ in the image?”, with balanced “Yes/No” answers for present and non-present objects. 
Finetuning and evaluation follow the protocol in POPE~\cite{POPE2023}.
We evaluate on the Popular Sampling split of POPE (3,000 samples). For training, we use MS COCO 2014 images with objects sampled from the 80 classes, selecting images with at least three objects and generating two “Yes” and two “No” questions per image ($\approx$160k pairs).

As demonstrated in Table~\ref{tab:pope_dataset_performance},
finetuning with LLL in addition to NTP increased the accuracy of  “Yes/No” answers on POPE dataset.
POPE accuracy measure was introduced in \cite{POPE2023} as a measure of hallucinations, where the higher accuracy means fewer hallucinations.
So, better accuracy coincides with a reduction in object hallucinations and, consequently, better performance on VQA tasks.

\begin{table}[t]
  \centering
  \scriptsize
  \setlength{\tabcolsep}{2pt}
  \renewcommand{\arraystretch}{0.92}
  \caption{
  Performance comparison on the POPE based on Popular sampling from MSCOCO images. 
  Higher accuracy indicates fewer hallucinated object predictions.
  RefCOCO $\rightarrow$ POPE denotes transfer: models are trained only on RefCOCO and evaluated on POPE.
  }
  \vspace{-2mm}
  \label{tab:pope_dataset_performance}
  \begin{tabular}{@{}lc@{}}
    \toprule
    \textbf{Method} & \textbf{Accuracy (\%)} \\
    \midrule
    \multicolumn{2}{@{}l}{\textit{LLaVA-v1.5-7B }} \\[0.2mm]
    \hspace{0.8em} Base Model                     & 86.23\% \\
    \hspace{0.8em} NTP                            & 90.03\% \\
    \hspace{0.8em} NTP + LLL                      & 92.40\% \\
    \midrule

    \multicolumn{2}{@{}l}{\textit{Qwen2.5-VL-7B-Instr.}} \\[0.2mm]
    \hspace{0.8em} Base Model                     & 86.77\% \\
    \hspace{0.8em} NTP                            & \second{90.50\%} \\
    \hspace{0.8em} NTP + LLL                      & \best{93.87\%} \\

    \multicolumn{2}{@{}l}{\textit{Qwen2.5-VL-7B-Instr. (RefCOCO $\rightarrow$ POPE)}} \\[0.2mm]
    \hspace{0.8em} NTP                            & 86.30\% \\
    \hspace{0.8em} NTP + LLL                      & 87.90\% \\
    \midrule

    \multicolumn{2}{@{}l}{\textit{SOTA}} \\[0.2mm]
    \hspace{0.8em} SPIN(LLaVA-v1.5-13B) \tcite{sarkar2025mitigating} & 88.83\% \\
    \hspace{0.8em} VCD(LLaVA-v1.5-13B)\tcite{leng2024mitigating} & 85.74\% \\ 
    \hspace{0.8em} SEA-LLaVA-7B ~\tcite{yin2025sea} & 87.60\% \\ 
    \bottomrule
  \end{tabular}
  \vspace{-2mm}
\end{table}

Importantly, \textbf{LLL improves accuracy even when the model is finetuned only on RefCOCO/+/g training splits (see Sec.~\ref{sec:RES}) and evaluated directly on POPE (RefCOCO $\rightarrow$ POPE)}, while NTP-only finetuning reduces accuracy relative to the base model, showing that LLL counteracts grounding degradation and enables cross-task generalization.

\vspace{-4mm}
\subsection{Logit Lens Object Confidence Scores}
\vspace{-2mm}



To qualitatively measure the impact of LLL on object localization, we consider the ratio 
of the average object bounding box Logit Lens score to the average score over the whole image as also used in \cite{choe2020evaluatingweaklysupervisedobject,jung2021betterexplanationsclassactivation}.
We call it the object confidence ratio.

\begin{figure}[t]
  \centering
    \centering
    \includegraphics[width=0.5\linewidth]{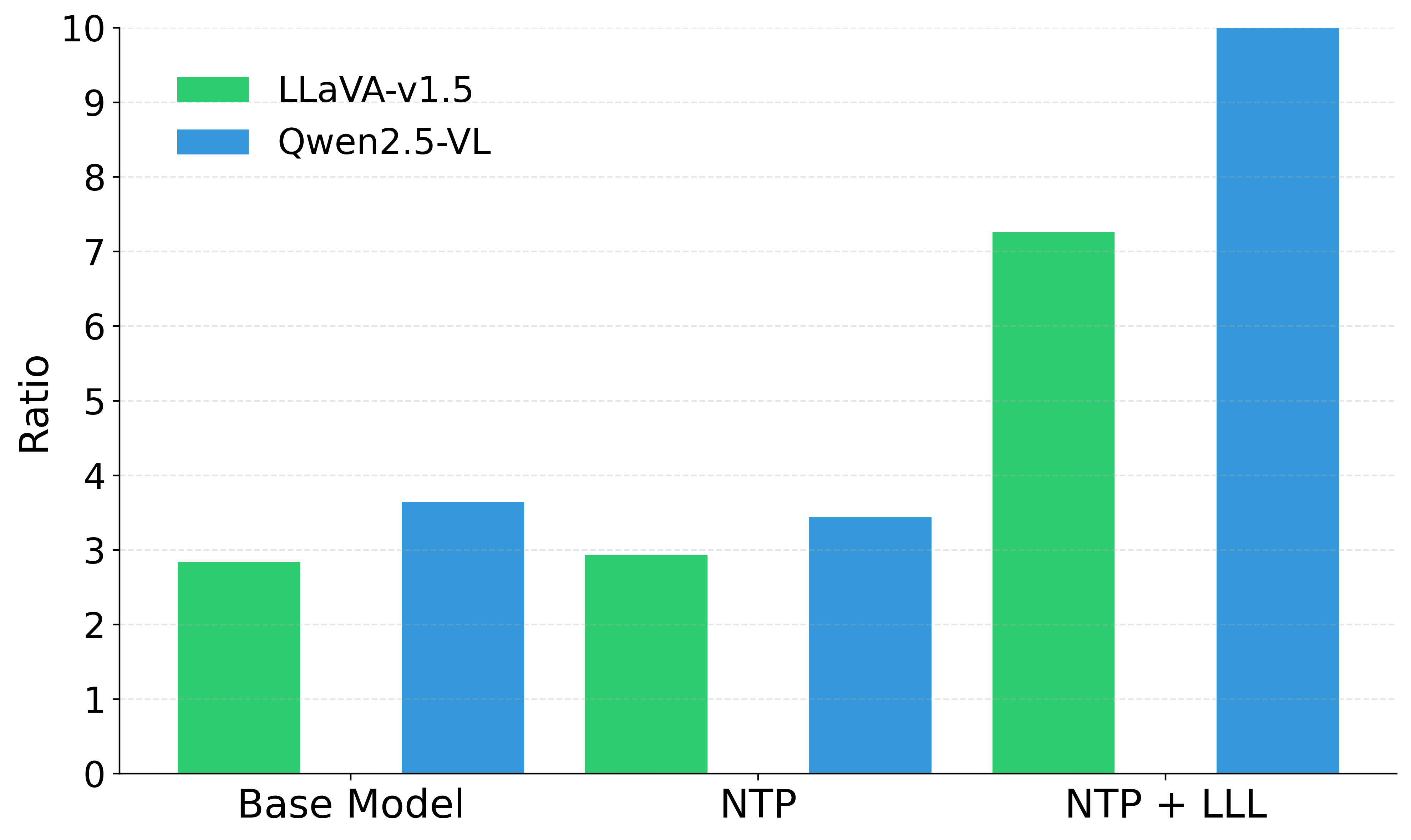}
  \vspace{-2mm}
  \caption{Bar plots illustrating the ratio of the object confidence scores of objects present in images within their bounding boxes to the confidence measure of the same objects averaged across all visual tokens in each image.}
  \label{fig:ratio_confidence}
  \vspace{-5mm}
\end{figure}

As demonstrated by the bar plots in Fig.~\ref{fig:ratio_confidence}, the addition of LLL to NTP leads to a threefold increase in object confidence score ratio.
This significant result confirms the huge effect of LLL on the object confidence maps shown in 
Fig.~\ref{fig:confidence_maps}. Moreover, the fact that NTP did not improve the ratio confirms that NTP provides only a weak supervisory signal for visual cues during fine-tuning. 
The results were obtained after finetuning on POPE.

\vspace{-4mm}
\subsection{Generalization to New Tasks}
\vspace{-2mm}
We further evaluate RES-trained models (Sec.~\ref{sec:RES}) on a pointing task for cross-task generalization and on Visual Question Answering for language capability assessment of VLMs.

\textbf{Zero-Shot Pointing to Object Centers.}
We show that the model finetuned with NTP + LLL is able to improve performance on a new localization task on a new set of images.
We constructed a subset of 1{,}500 images from PixMo-Points~\cite{deitke2025molmo} for evaluation. We randomly sample from 150 object categories with the highest occurrence frequency. 
We stress that the images in PixMo-Points dataset have no known overlap with MS COCO, and that the classes in the constructed subset do not overlap with those used during training. 
Each sample provides a short textual prompt and a human-annotated point marking the object’s center. The model must output the corresponding $(x, y)$ coordinates of the referred object’s center in the pixel space.

Table~\ref{tab:pixmo_dataset_performance} presents zero-shot evaluation results on the PixMo-Points subset. 
All evaluations are performed in pixel space, with coordinates parsed from the model’s textual output using a fixed prompt template. We report the median distance in pixels. 
Incorporating LLL during fine-tuning substantially improves both models, reducing median localization error.
Notably, Qwen2.5-VL with LLL achieves a lower median distance than several proprietary and task-specific baselines, and it outperforms Molmo-7B, which was trained with explicit coordinate-level supervision on PixMo. These results indicate that LLL enables strong transfer of spatial grounding learned from RefCOCO/+/g data to previously unseen object categories in PixMo.

\begin{table}[hbtp]
  \centering
  \scriptsize
  \setlength{\tabcolsep}{2pt}
  \renewcommand{\arraystretch}{0.92}

  \caption{Zero-shot performance on the Pixmo. LLaVA-v1.5 and Qwen2.5-VL were finetuned on RefCOCO training set.}
  \vspace{-2mm}
  \label{tab:pixmo_dataset_performance}

  \begin{tabular}{@{}lccc@{}}
    \toprule
    \textbf{Model} & \textbf{Base Model} & \textbf{NTP} & \textbf{NTP + LLL} \\
    \midrule
    LLaVA-v1.5-7B & 13.60 & 12.95 & 9.21 \\
    Qwen2.5-VL-7B-Instr. & 7.07 & \second{6.95} & \best{6.30} \\
    \midrule
    \multicolumn{4}{@{}l}{\textit{SOTA}} \\[0.2mm]
    Molmo-7B-D~\tcite{deitke2025molmo} & 7.13 & -- & -- \\
    GPT-4o & 7.07 & -- & -- \\
    Gemini-2.0 Flash & 7.21 & -- & -- \\
    \bottomrule
  \end{tabular}

  \vspace{-5mm}
\end{table}

\textbf{Visual Question Answering.} We evaluate the impact of LLL on standard VQA benchmarks, VQAv2 and GQA (Table~\ref{tab:vqa_dataset_performance}). Unlike methods such as SEA-LLaVA, which use a two-stage training pipeline with an explicit vision–language alignment stage during adapter training, our approach uses a single-stage finetuning procedure that jointly updates both the projection module and LLM layers. Despite this simpler setup, our method achieves performance comparable to SEA-LLaVA on VQAv2, bringing LLaVA-v1.5 close to their results. On VQAv2, we observe consistent improvements over NTP-only finetuning, with modest gains for LLaVA and around +0.4 for Qwen2.5-VL, showing that LLL does not hurt language performance.

On the more challenging GQA benchmark, which requires reasoning and stronger grounding, we observe that NTP alone slightly degrades performance for LLaVA, whereas NTP+LLL recovers and improves slightly over the base model. For Qwen2.5-VL, the improvement over NTP is more pronounced, showing that LLL helps the model better use visual information. Overall, NTP+LLL leads to consistent, though modest, improvements on VQA tasks while significantly improving grounding.

It is also important to note that for instruction-tuning data without localization annotations (e.g., LLaVA-150K), our training reduces to standard NTP during training. Therefore, the observed gains are achieved despite limited grounding supervision in these samples, suggesting that further improvements could be achieved by incorporating weak supervision in the form of object masks generated by segmentation models such as SAM and then applied through LLL.

More experimental results, ablations, 
visualizations, and additional implementation details are provided in the appendix.

\begin{table}[hbtp]
  \centering
  \tiny
  \setlength{\tabcolsep}{2.5pt}
  \renewcommand{\arraystretch}{0.92}
  \caption{Comparison on visual question answering benchmarks.}
  \vspace{-2mm}
  \label{tab:vqa_dataset_performance}
  \begin{tabular}{@{}lcccccccc@{}}
    \toprule
    \textbf{Benchmark}
    & \multicolumn{3}{c}{\textbf{LLaVA-v1.5-7B}}
    & \multicolumn{3}{c}{\textbf{Qwen2.5-VL-7B}}
    & \multicolumn{2}{c}{\textbf{SOTA}} \\
    \cmidrule(lr){2-4}
    \cmidrule(lr){5-7}
    \cmidrule(l){8-9}
    & \textbf{Base} & \textbf{NTP} & \textbf{NTP+LLL}
    & \textbf{Base} & \textbf{NTP} & \textbf{NTP+LLL}
    & \textbf{SEA-LLaVA-7B} & \textbf{InternVL2-7B} \\
    \midrule
    VQAv2
    & 78.8 & 78.9 & 79.0
    & 80.5 & 80.6 & \best{80.9}
    & \second{79.1} & 77.5 \\
    GQA
    & 62.0 & 61.9 & \second{62.1}
    & 60.4 & 60.5 & 60.7
    & \best{63.2} & 61.3 \\
    \bottomrule
  \end{tabular}
  \vspace{-10mm}
\end{table}
\section{Related Work}
\vspace{-2mm}


Our work is focused on the dominant group of VLMs, which are
instruction-tuned, autoregressive VLMs~\cite{li2023blip, liu2023visual, bai2023qwenvlversatilevisionlanguagemodel, zhu2023minigpt, lu2024deepseek}.
They achieve strong results on captioning~\cite{li2023blip, wang2021simvlm}, retrieval~\cite{radford2021learning,yu2022coca}, and VQA~\cite{alayrac2022flamingo}.  
However, scaling alone does not guarantee faithful visual grounding: VLMs hallucinate plausible details without images~\cite{payattentionimageECCV2024}, can ignore large fractions of visual tokens~\cite{chen2024imageworth12tokens}, and suffer from ``attention sinks’’~\cite{kang2025attentionsink}. Existing benchmarks also insufficiently probe vision-centric reasoning~\cite{tong2024cambrian, tong2024eyes}. These issues suggest that next-token prediction (NTP) provides weak patch-level supervision, allowing visual embeddings to drift away from their original image content during LLM processing.

A major direction for improving grounding is equipping VLMs with localization heads and finetuning on grounding datasets. LISA~\cite{lai2024lisa} introduces a \texttt{[SEG]} token and mask decoder; F-LMM~\cite{wu2025f} refines attention-derived masks using a lightweight module; and GLAMM~\cite{rasheed2024glamm} adds a grounding encoder and pixel decoder. While effective, these systems predict boxes or masks externally and do not bind language concepts to the visual tokens inside the LLM, which is our main contribution.

Logit Lens~\cite{logitlens2020, LogitLensICLR2025} projects intermediate VLM embeddings into the vocabulary space, revealing which textual concepts are represented at each layer. Prior work shows that these projections expose a drift of visual embeddings toward language-dominated representations during modality mixing~\cite{payattentionimageECCV2024, chen2024imageworth12tokens, wu2025f}. Moreover, removing visual tokens inside ground-truth regions strongly degrades performance, indicating that these patches contain key semantic information that may become difficult to recover after such drift~\cite{neo2025interpretingvisualinformationprocessing}. Existing approaches often address grounding through coordinate-based models, such as Pix2Seq and Kosmos-2~\cite{pix2seqICLR2022,kosmos2-ICLR2023}, or through global alignment models such as CLIP, UNITER, and ViLBERT~\cite{radford2021learning, chen2020uniter, lu2019vilbert}.

Recent token-level methods improve alignment with semantic labels or external vision models, such as SEA and VIRAL~\cite{yoon2025visual}. In contrast, we directly align visual tokens with textual concepts in the logit space, preserving grounding inside the LLM without architectural changes.

Visualization methods are widely used to analyze how visual encoders and VLMs interpret images. 
Attention-based analyses \cite{darcet2023vision, kang2025attentionsink,payattentionimageECCV2024} and tools such as VL-InterpreT \cite{aflalo2022vlinterpretinteractivevisualizationtool} examine token-level behavior, while Grad-CAM \cite{Selvaraju_2019} provides gradient-based spatial cues.


\vspace{-6mm}
\section{Conclusion}
\vspace{-3mm}




We introduced Logit Lens Loss (LLL), a lightweight auxiliary objective for improving patch-level explanations in autoregressive Vision-Language Models. LLL directly aligns visual-token embeddings with vocabulary concepts describing their image regions, without adding a mask decoder, localization head, or other architectural module.
From an XAI perspective, LLL makes Logit Lens maps more faithful by deriving explanations from the VLM's own visual-token embeddings and vocabulary projection. Across LLaVA-v1.5-7B and Qwen2.5-VL-7B, LLL produces sharper object confidence maps, improves grounding and hallucination metrics, transfers to zero-shot pointing, and preserves 
VQA performance. These results suggest that better internal grounding can make foundation VLMs both more interpretable and more effective.


\vspace{-3mm}
\section*{Acknowledgements}
We thank Ying Jin for help with this work. This work was supported by the NSF award IIS-2331768.
%
%
\bibliographystyle{splncs04}
\bibliography{references}

\begin{thebibliography}{10}
\providecommand{\url}[1]{\texttt{#1}}
\providecommand{\urlprefix}{URL }
\providecommand{\doi}[1]{https://doi.org/#1}

\bibitem{aflalo2022vlinterpretinteractivevisualizationtool}
Aflalo, E., Du, M., Tseng, S.Y., Liu, Y., Wu, C., Duan, N., Lal, V.: Vl-interpret: An interactive visualization tool for interpreting vision-language transformers (2022), \url{https://arxiv.org/abs/2203.17247}

\bibitem{alayrac2022flamingo}
Alayrac, J.B., Donahue, J., Luc, P., Miech, A., Barr, I., Hasson, Y., Lenc, K., Mensch, A., Millican, K., Reynolds, M., et~al.: Flamingo: a visual language model for few-shot learning. In: Advances in Neural Information Processing Systems. vol.~35, pp. 23716--23736 (2022)

\bibitem{bai2023qwenvlversatilevisionlanguagemodel}
Bai, J., Bai, S., Yang, S., Wang, S., Tan, S., Wang, P., Lin, J., Zhou, C., Zhou, J.: Qwen-vl: A versatile vision-language model for understanding, localization, text reading, and beyond (2023), \url{https://arxiv.org/abs/2308.12966}

\bibitem{bai2025qwen2}
Bai, S., Chen, K., Liu, X., Wang, J., Ge, W., Song, S., Dang, K., Wang, P., Wang, S., Tang, J., et~al.: Qwen2. 5-vl technical report. arXiv preprint arXiv:2502.13923  (2025)

\bibitem{chen2024imageworth12tokens}
Chen, L., Zhao, H., Liu, T., Bai, S., Lin, J., Zhou, C., Chang, B.: An image is worth 1/2 tokens after layer 2: Plug-and-play inference acceleration for large vision-language models (2024)

\bibitem{pix2seqICLR2022}
Chen, T., Saxena, S., Li, L., Fleet, D.J., Hinton, G.: Pix2seq: A language modeling framework for object detection. In: ICLR (2022)

\bibitem{chen2020uniter}
Chen, Y.C., Li, L., Yu, L., El~Kholy, A., Ahmed, F., Gan, Z., Cheng, Y., Liu, J.: Uniter: Universal image-text representation learning. In: European conference on computer vision. pp. 104--120. Springer (2020)

\bibitem{choe2020evaluatingweaklysupervisedobject}
Choe, J., Oh, S.J., Lee, S., Chun, S., Akata, Z., Shim, H.: Evaluating weakly supervised object localization methods right (2020), \url{https://arxiv.org/abs/2001.07437}

\bibitem{darcet2023vision}
Darcet, T., Oquab, M., Mairal, J., Bojanowski, P.: Vision transformers need registers. arXiv preprint arXiv:2309.16588  (2023)

\bibitem{deitke2025molmo}
Deitke, M., Clark, C., Lee, S., Tripathi, R., Yang, Y., Park, J.S., Salehi, M., Muennighoff, N., Lo, K., Soldaini, L., et~al.: Molmo and pixmo: Open weights and open data for state-of-the-art vision-language models. In: Proceedings of the Computer Vision and Pattern Recognition Conference. pp. 91--104 (2025)

\bibitem{Esmaeilkhani_2026_WACV}
Esmaeilkhani, P., Latecki, L.J.: Direct visual grounding by directing attention of visual tokens. In: Proceedings of the IEEE/CVF Winter Conference on Applications of Computer Vision (WACV). pp. 5787--5797 (March 2026)

\bibitem{fu2025hidden}
Fu, S., Bonnen, T., Guillory, D., Darrell, T.: Hidden in plain sight: Vlms overlook their visual representations. arXiv preprint arXiv:2506.08008  (2025)

\bibitem{LogitLensICLR2025}
Jiang, N., Kachinthaya, A., Petryk, S., Gandelsman, Y.: Interpreting and editing vision-language representations to mitigate hallucinations. In: ICLR (2025)

\bibitem{jung2021betterexplanationsclassactivation}
Jung, H., Oh, Y.: Towards better explanations of class activation mapping (2021), \url{https://arxiv.org/abs/2102.05228}

\bibitem{whatsimagedeepdCVPR2025}
Kaduri, O., Bagon, S., Dekel, T.: What's in the image? a deep-dive into the vision of vision language models. In: CVPR (2025)

\bibitem{kang2025attentionsink}
Kang, S., Kim, J., Kim, J., Hwang, S.J.: See what you are told: Visual attention sink in large multimodal models. In: ICLR (2025)

\bibitem{SAM2023}
Kirillov, A., Mintun, E., Ravi, N., Mao, H., Rolland, L., Schmidt, D., Li, Z., Doll{\'a}r, P., Girshick, R.: Segment anything. In: ICCV (2023)

\bibitem{lai2024lisa}
Lai, X., Tian, Z., Chen, Y., Li, Y., Yuan, Y., Liu, S., Jia, J.: Lisa: Reasoning segmentation via large language model. In: Proceedings of the IEEE/CVF Conference on Computer Vision and Pattern Recognition. pp. 9579--9589 (2024)

\bibitem{leng2024mitigating}
Leng, S., Zhang, H., Chen, G., Li, X., Lu, S., Miao, C., Bing, L.: Mitigating object hallucinations in large vision-language models through visual contrastive decoding. In: Proceedings of the IEEE/CVF Conference on Computer Vision and Pattern Recognition. pp. 13872--13882 (2024)

\bibitem{li2023blip}
Li, J., Li, D., Savarese, S., Hoi, S.: Blip-2: Bootstrapping language-image pre-training with frozen image encoders and large language models. In: International conference on machine learning. pp. 19730--19742. PMLR (2023)

\bibitem{POPE2023}
Li, Y., Du, Y., Zhou, K., Wang, J., Zhao, W.X., Wen, J.R.: Evaluating object hallucination in large vision-language models. In: Conference on Empirical Methods in Natural Language Processing (EMNLP) (2023)

\bibitem{liu2023visual}
Liu, H., Li, C., Wu, Q., Lee, Y.J.: Visual instruction tuning. Advances in neural information processing systems  \textbf{36},  34892--34916 (2023)

\bibitem{payattentionimageECCV2024}
Liu, S., Zheng, K., Chen, W.: Paying more attention to image: A training-free method for alleviating hallucination in lvlms. In: European Conference on Computer Vision. pp. 125--140. Springer (2024)

\bibitem{lu2024deepseek}
Lu, H., Liu, W., Zhang, B., Wang, B., Dong, K., Liu, B., Sun, J., Ren, T., Li, Z., Yang, H., et~al.: Deepseek-vl: towards real-world vision-language understanding. arXiv preprint arXiv:2403.05525  (2024)

\bibitem{lu2019vilbert}
Lu, J., Batra, D., Parikh, D., Lee, S.: Vilbert: Pretraining task-agnostic visiolinguistic representations for vision-and-language tasks. Advances in neural information processing systems  \textbf{32} (2019)

\bibitem{mikolov2013distributedrepresentationswordsphrases}
Mikolov, T., Sutskever, I., Chen, K., Corrado, G., Dean, J.: Distributed representations of words and phrases and their compositionality (2013), \url{https://arxiv.org/abs/1310.4546}

\bibitem{neo2025interpretingvisualinformationprocessing}
Neo, C., Ong, L., Torr, P., Geva, M., Krueger, D., Barez, F.: Towards interpreting visual information processing in vision-language models (2025), \url{https://arxiv.org/abs/2410.07149}

\bibitem{logitlens2020}
nostalgebraist: Interpreting gpt: The logit lens. \url{https://www.lesswrong.com/posts/AcKRB8wDpdaN6v6ru/interpreting-gpt-the-logit-lens} (aug 2020), lessWrong

\bibitem{oquab2024dinov2learningrobustvisual}
Oquab, M., Darcet, T., Moutakanni, T., Vo, H., Szafraniec, M., Khalidov, V., Fernandez, P., Haziza, D., Massa, F., El-Nouby, A., Assran, M., Ballas, N., Galuba, W., Howes, R., Huang, P.Y., Li, S.W., Misra, I., Rabbat, M., Sharma, V., Synnaeve, G., Xu, H., Jegou, H., Mairal, J., Labatut, P., Joulin, A., Bojanowski, P.: Dinov2: Learning robust visual features without supervision (2024), \url{https://arxiv.org/abs/2304.07193}

\bibitem{kosmos2-ICLR2023}
Peng, Z., Wang, W., Dong, L., Hao, Y., Huang, S., Ma, S., Wei, F.: Kosmos-2: Grounding multimodal large language models to the world. In: ICLR (2023)

\bibitem{radford2021learning}
Radford, A., Kim, J.W., Hallacy, C., Ramesh, A., Goh, G., Agarwal, S., Sastry, G., Askell, A., Mishkin, P., Clark, J., et~al.: Learning transferable visual models from natural language supervision. In: International conference on machine learning. pp. 8748--8763. PmLR (2021)

\bibitem{rasheed2024glamm}
Rasheed, H., Maaz, M., Shaji, S., Shaker, A., Khan, S., Cholakkal, H., Anwer, R.M., Xing, E., Yang, M.H., Khan, F.S.: Glamm: Pixel grounding large multimodal model. In: Proceedings of the IEEE/CVF Conference on Computer Vision and Pattern Recognition. pp. 13009--13018 (2024)

\bibitem{sarkar2025mitigating}
Sarkar, S., Che, Y., Gavin, A., Beerel, P.A., Kundu, S.: Mitigating hallucinations in vision-language models through image-guided head suppression. In: Proceedings of the 2025 Conference on Empirical Methods in Natural Language Processing. pp. 12492--12511 (2025)

\bibitem{Selvaraju_2019}
Selvaraju, R.R., Cogswell, M., Das, A., Vedantam, R., Parikh, D., Batra, D.: Grad-cam: Visual explanations from deep networks via gradient-based localization. International Journal of Computer Vision  \textbf{128}(2),  336–359 (Oct 2019). \doi{10.1007/s11263-019-01228-7}, \url{http://dx.doi.org/10.1007/s11263-019-01228-7}

\bibitem{tong2024cambrian}
Tong, P., Brown, E., Wu, P., Woo, S., IYER, A.J.V., Akula, S.C., Yang, S., Yang, J., Middepogu, M., Wang, Z., et~al.: Cambrian-1: A fully open, vision-centric exploration of multimodal llms. Advances in Neural Information Processing Systems  \textbf{37},  87310--87356 (2024)

\bibitem{tong2024eyes}
Tong, S., Liu, Z., Zhai, Y., Ma, Y., LeCun, Y., Xie, S.: Eyes wide shut? exploring the visual shortcomings of multimodal llms. In: Proceedings of the IEEE/CVF Conference on Computer Vision and Pattern Recognition. pp. 9568--9578 (2024)

\bibitem{wang2024mllm}
Wang, C., Chen, X., Zhang, N., Tian, B., Xu, H., Deng, S., Chen, H.: Mllm can see? dynamic correction decoding for hallucination mitigation. arXiv preprint arXiv:2410.11779  (2024)

\bibitem{wang2021simvlm}
Wang, Z., Yu, J., Yu, A.W., Dai, Z., Tsvetkov, Y., Cao, Y.: Simvlm: Simple visual language model pretraining with weak supervision. arXiv preprint arXiv:2108.10904  (2021)

\bibitem{wu2025flmmgroundingfrozenlarge}
Wu, S., Jin, S., Zhang, W., Xu, L., Liu, W., Li, W., Loy, C.C.: F-lmm: Grounding frozen large multimodal models (2025), \url{https://arxiv.org/abs/2406.05821}

\bibitem{wu2025f}
Wu, S., Jin, S., Zhang, W., Xu, L., Liu, W., Li, W., Loy, C.C.: F-lmm: Grounding frozen large multimodal models. In: Proceedings of the Computer Vision and Pattern Recognition Conference. pp. 24710--24721 (2025)

\bibitem{xia2024gsvageneralizedsegmentationmultimodal}
Xia, Z., Han, D., Han, Y., Pan, X., Song, S., Huang, G.: Gsva: Generalized segmentation via multimodal large language models (2024), \url{https://arxiv.org/abs/2312.10103}

\bibitem{xiao2024towards}
Xiao, L., Yang, X., Lan, X., Wang, Y., Xu, C.: Towards visual grounding: A survey. arXiv preprint arXiv:2412.20206  (2024)

\bibitem{yin2025sea}
Yin, Y., Zhao, Y., Zhang, Y., Zhang, Y., Lin, K., Wang, J., Tao, X., Wan, P., Zhang, W., Zhao, F.: Sea: Supervised embedding alignment for token-level visual-textual integration in mllms. In: Proceedings of the 2025 Conference on Empirical Methods in Natural Language Processing. pp. 1058--1070 (2025)

\bibitem{yoon2025visual}
Yoon, H., Jung, J., Kim, J., Choi, H., Shin, H., Lim, S., An, H., Kim, C., Han, J., Kim, D., et~al.: Visual representation alignment for multimodal large language models. arXiv preprint arXiv:2509.07979  (2025)

\bibitem{yu2022coca}
Yu, J., Wang, Z., Vasudevan, V., Yeung, L., Seyedhosseini, M., Wu, Y.: Coca: Contrastive captioners are image-text foundation models. arXiv preprint arXiv:2205.01917  (2022)

\bibitem{yu2016modeling}
Yu, L., Poirson, P., Yang, S., Berg, A.C., Berg, T.L.: Modeling context in referring expressions. In: European conference on computer vision. pp. 69--85. Springer (2016)

\bibitem{zhang2024psalm}
Zhang, Z., Ma, Y., Zhang, E., Bai, X.: Psalm: Pixelwise segmentation with large multi-modal model. In: European Conference on Computer Vision. pp. 74--91. Springer (2024)

\bibitem{zhu2023minigpt}
Zhu, D., Chen, J., Shen, X., Li, X., Elhoseiny, M.: Minigpt-4: Enhancing vision-language understanding with advanced large language models. arXiv preprint arXiv:2304.10592  (2023)

\end{thebibliography}

\end{document}


\graphicspath{{./figures/}}

\title{Supplementary Material:\\
Logit Lens Supervision for Patch-Level Explanations in Vision-Language Models}
\titlerunning{Supplementary Material}


\author{
Parsa Esmaeilkhani\inst{1} \and
Longin Jan Latecki\inst{1}
}

\authorrunning{P.~Esmaeilkhani and L.~J.~Latecki}

\institute{
Department of Computer and Information Sciences, Temple University,\\
Philadelphia, PA, USA\\
\email{\{parsa.esmaeilkhani, latecki\}@temple.edu}
}
\maketitle

\section{Implementation Configuration}
\label{sec:supp-implementation}

We finetuned LLaVA-v1.5-7B and Qwen2.5-VL-7B-Instruct using DeepSpeed on four NVIDIA A100 GPUs with 80 GB VRAM each. The vision encoder remained frozen except for a trainable projection layer, while Low-Rank Adaptation (LoRA) was applied to the language-model layers. Both LLaVA-v1.5 and Qwen2.5-VL used LoRA rank 16. For the Logit Lens Loss (LLL), the language head, which corresponds to the unembedding matrix, was frozen for consistency. Training used a per-GPU batch size of 16 and gradient accumulation of 4, resulting in an effective global batch size of 256. Optimization followed a cosine-decay learning-rate schedule with initial learning rate $5\times10^{-5}$ and a linear warm-up over the first 3\% of total training steps. For all tasks except zero-shot pointing, we finetuned the models for one epoch on the corresponding training sets.

For the commercial VLM baselines, GPT-4o and Gemini-2.0 Flash, we queried the public APIs using their default system prompts. We used greedy decoding when supported, did not apply top-$k$ or top-$p$ sampling, and left the temperature parameter at its default setting, ensuring that outputs were directly comparable to those from our finetuned open-source models.

\paragraph{Training prompt.}
We use the same prompt template for training and evaluation, instructing the VLM to output a single bounding box for the referred object:
\begin{quote}\small
\textbf{Given the image and the referring expression, output the bounding box coordinates of the referred object in pixel coordinates.}
\end{quote}

\paragraph{Logit Lens confidence map-to-mask generation and thresholding.}
For heatmap thresholding, we binarize the heatmap using a scalar threshold
$\tau$ computed from per-image heatmap statistics. We evaluate
\[
\tau \in \{\mu-\sigma,\ \mu,\ \mu+\sigma\},
\]
where $\mu$ and $\sigma$ are the mean and standard deviation of heatmap values
for the given image. Pixels with values above $\tau$ are retained in the binary
mask. For Logit Lens maps, $\tau=\mu$ performs best.

\section{Ablations}
\label{sec:supp-ablations}

We first evaluate how the layer choice for applying LLL affects performance. As shown in \cref{tab:supp-ablation-layers}, applying LLL to the last layer only achieves the best accuracy, 93.87\%, outperforming both top-layer aggregation and middle-layer aggregation. This is expected because the final layer is most closely aligned with the language head, making the LLL constraint more effective.

\Cref{tab:supp-lambda-ablation} shows the influence of the weighting coefficient $\lambda$ in the total loss. A moderate weighting provides the strongest gains: accuracy peaks at $\lambda=0.50$, while both smaller and larger values lead to slight declines.

\begin{table}[hbtp]
  \centering
  \scriptsize
  \setlength{\tabcolsep}{3pt}
  \renewcommand{\arraystretch}{0.92}
  \caption{Ablation of layer choices for Logit Lens Loss on the POPE dataset for Qwen2.5-VL-7B-Instruct with NTP+LLL.}
  \label{tab:supp-ablation-layers}
  \begin{tabular}{lc}
    \toprule
    \textbf{Layer} & \textbf{Accuracy (\%)} \\
    \midrule
    Top 4 layers            & 90.86 \\
    Middle layers (13--16)  & 91.37 \\
    Last layer only         & \textbf{93.87} \\
    \bottomrule
  \end{tabular}
\end{table}

\begin{table}[t]
  \centering
  \scriptsize
  \setlength{\tabcolsep}{3pt}
  \renewcommand{\arraystretch}{0.92}
  \caption{Effect of varying the hyperparameter $\lambda$ in the combined loss $\mathcal{L}_{\mathrm{total}}=\mathcal{L}_{\mathrm{NTP}}+\lambda\mathcal{L}_{\mathrm{LLL}}$ on POPE for Qwen2.5-VL-7B-Instruct.}
  \label{tab:supp-lambda-ablation}
  \begin{tabular}{lc}
    \toprule
    \textbf{$\lambda$} & \textbf{Accuracy (\%)} \\
    \midrule
    0.25 & 92.41 \\
    0.50 & \textbf{93.87} \\
    0.75 & 93.12 \\
    1.00 & 92.78 \\
    \bottomrule
  \end{tabular}
\end{table}

\section{Referring Expression Comprehension}
\label{sec:supp-rec}

\Cref{tab:supp-refcoco-rec} compares our approach with existing state-of-the-art models on RefCOCO, RefCOCO+, and RefCOCOg for Referring Expression Comprehension (REC). Given a natural-language phrase describing an object, the model predicts the bounding box of the queried object as text output. Finetuning with LLL consistently improves performance over both the base and NTP-finetuned models across all datasets. Qwen2.5-VL with LLL achieves results that rival or surpass several SOTA VLMs while approaching the performance of substantially larger models such as CogVLM-17B. These findings demonstrate that LLL provides competitive grounding capabilities without architectural changes, highlighting its effectiveness as a lightweight auxiliary objective for enhancing visual alignment in VLMs.

\begin{table}[t]
  \centering
  \scriptsize
  \setlength{\tabcolsep}{2pt}
  \renewcommand{\arraystretch}{0.92}
  \caption{Comparison on the REC task measured by IoU@0.5. SOTA methods are finetuning-based VLMs.}
  \label{tab:supp-refcoco-rec}
  \begin{tabular}{@{}lccc@{}}
    \toprule
    \textbf{Method} & \textbf{RefCOCO} & \textbf{RefCOCO+} & \textbf{RefCOCOg} \\
    \midrule
    \multicolumn{4}{@{}l}{\textit{LLaVA-v1.5-7B}} \\[0.2mm]
    \hspace{0.8em} Base Model & 52.90 & 42.55 & 45.00 \\
    \hspace{0.8em} NTP & 53.20 & 42.90 & 45.40 \\
    \hspace{0.8em} NTP+LLL & 53.95 & 43.40 & 46.00 \\
    \midrule
    \multicolumn{4}{@{}l}{\textit{Qwen2.5-VL-7B-Instruct}} \\[0.2mm]
    \hspace{0.8em} Base Model & 90.05 & 85.40 & 86.60 \\
    \hspace{0.8em} NTP & 90.60 & 85.70 & 86.80 \\
    \hspace{0.8em} NTP+LLL & \second{91.50} & \second{86.40} & \second{87.30} \\
    \midrule
    \multicolumn{4}{@{}l}{\textit{SOTA}} \\[0.2mm]
    \hspace{0.8em} LLaVA-G-7B \tcite{zhang2023llavagroundinggroundedvisualchat} & 89.16 & 81.68 & 84.82 \\
    \hspace{0.8em} Shikra-7B \tcite{chen2023shikraunleashingmultimodalllms} & 87.00 & 81.60 & 82.30 \\
    \hspace{0.8em} Ferret-7B \tcite{you2023ferretrefergroundgranularity} & 87.50 & 80.80 & 83.90 \\
    \hspace{0.8em} Shikra-13B \tcite{chen2023shikraunleashingmultimodalllms} & 87.80 & 82.90 & 82.60 \\
    \hspace{0.8em} Ferret-13B \tcite{you2023ferretrefergroundgranularity} & 89.50 & 82.80 & 85.80 \\
    \hspace{0.8em} CogVLM-17B \tcite{wang2024cogvlmvisualexpertpretrained} & \best{92.80} & \best{88.70} & \best{89.80} \\
    \bottomrule
  \end{tabular}
\end{table}

\section{Detailed Analysis of the Weak NTP Gradient Signal}
The NTP loss is evaluated only at supervised text-output positions. Consequently,
the final-layer visual representations do not participate directly in next-token
prediction, and therefore
\[
\nabla_{h_L(s)} L_{\mathrm{NTP}} = 0
\]
for every visual-token position $s$. Earlier visual representations
$h_l(s)$, with $l<L$, may nevertheless receive an indirect gradient because
they influence final-layer text representations through causal self-attention
in subsequent decoder blocks. However, text tokens often assign only weak
attention to most visual tokens.

Let $\mathcal{T}_{\mathrm{sup}}$ denote the set of supervised text-output
positions at which the NTP loss is evaluated. By the chain rule, for a visual
token $s$ at layer $l<L$,
\begin{equation}
\label{eq:gradofNTP}
\nabla_{h_l(s)} L_{\mathrm{NTP}}(\theta)
=
\sum_{j \in \mathcal{T}_{\mathrm{sup}}}
\left(
\frac{\partial h_L(j)}
     {\partial h_l(s)}
\right)^{\!\top}
\underbrace{
\nabla_{h_L(j)} L_{\mathrm{NTP}}(\theta)
}_{\text{local gradient at a supervised text position}} .
\end{equation}
Thus, NTP does not provide a direct gradient to the final-layer visual
representation $h_L(s)$. Instead, its gradient reaches an earlier visual
representation $h_l(s)$ only through the computational paths by which that
representation affects supervised text positions in layers $l+1,\ldots,L$.

In particular, for $l=L-1$, the cross-position dependence of a final-layer
text representation $h_L(j)$ on a visual representation $h_{L-1}(s)$ arises
through the causal self-attention sublayer of the final decoder block. Assuming
bounded hidden-state norms and bounded operator norms of the transformations
within this block, the corresponding Jacobian can be bounded as
\begin{equation}
\label{eq:att_jacobian}
\left\|
\frac{\partial h_L(j)}
     {\partial h_{L-1}(s)}
\right\|
\le
C_{\mathrm{att}}\,a^{(L)}_{js},
\end{equation}
where $a^{(L)}_{js}$ is the attention weight in the final decoder block from
text position $j$ to visual position $s$. The constant $C_{\mathrm{att}}$
absorbs the bounded contributions of the query, key, value, and output
projections $(W_Q,W_K,W_V,W_O)$, layer normalization, and the subsequent
position-wise transformations within the block.

Substituting Eq.~\eqref{eq:att_jacobian} into
Eq.~\eqref{eq:gradofNTP} and applying the triangle inequality gives
\begin{equation}
\label{eq:final_bound}
\left\|
\nabla_{h_{L-1}(s)} L_{\mathrm{NTP}}
\right\|
\le
C_{\mathrm{att}}
\sum_{j\in\mathcal{T}_{\mathrm{sup}}}
\left\|
\nabla_{h_L(j)} L_{\mathrm{NTP}}
\right\|
a^{(L)}_{js}.
\end{equation}

For a per-token cross-entropy loss
$\ell_j=-\log p_\theta(y_j\mid h_L(j))$, the local gradient is
\[
\nabla_{h_L(j)} \ell_j
=
U^\top\!\left(p_j-e_{y_j}\right),
\]
and hence
\[
\left\|
\nabla_{h_L(j)} \ell_j
\right\|
\le
\sqrt{2}\,\|U\|_2,
\]
because $\|p_j-e_{y_j}\|_2\le\sqrt{2}$. If
$L_{\mathrm{NTP}}$ is averaged over
$|\mathcal{T}_{\mathrm{sup}}|$ supervised positions, the corresponding bound
contains the additional factor
$1/|\mathcal{T}_{\mathrm{sup}}|$.

Equation~\eqref{eq:final_bound} therefore shows that the indirect NTP gradient
reaching $h_{L-1}(s)$ is controlled by the attention weights
$a^{(L)}_{js}$. When these weights are small for a visual token $s$, the
corresponding supervisory signal can be strongly attenuated
\cite{Esmaeilkhani_2026_WACV,chen2024imageworth12tokens}. In contrast, LLL
provides a direct gradient to the final-layer visual representation $h_L(s)$
through the unembedding matrix.

\section{Object Confidence Map Limitation}
\label{sec:supp-limitations}

\begin{figure*}[t]
  \centering
  \begin{subfigure}[t]{0.24\textwidth}
    \vspace{0pt}
    \includegraphics[width=\textwidth]{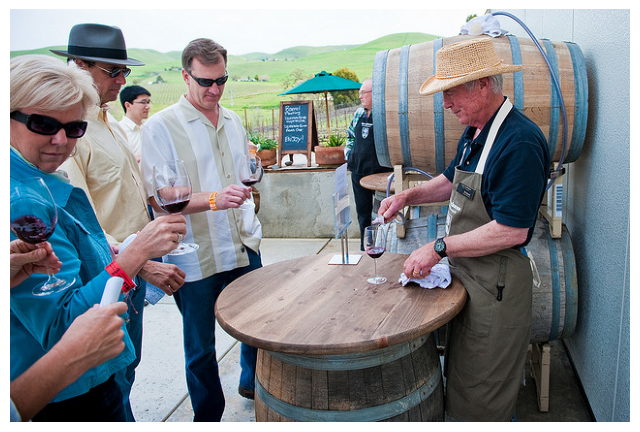}
    \caption{Object: person}
  \end{subfigure}\hfill
  \begin{subfigure}[t]{0.24\textwidth}
    \vspace{0pt}
    \includegraphics[width=\textwidth]{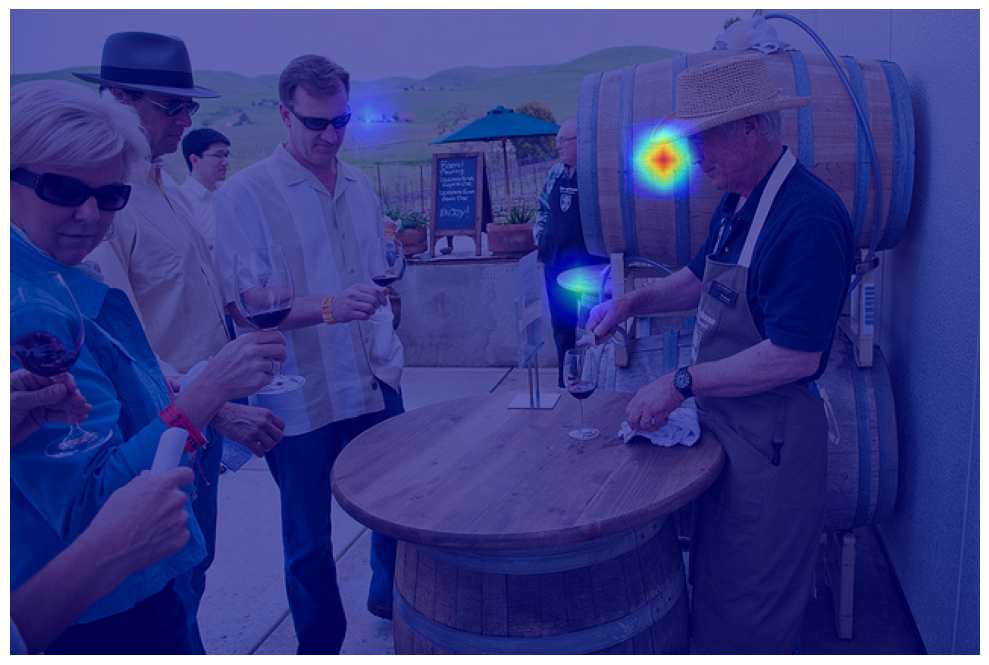}
    \caption{Base}
  \end{subfigure}\hfill
  \begin{subfigure}[t]{0.24\textwidth}
    \vspace{0pt}
    \includegraphics[width=\textwidth]{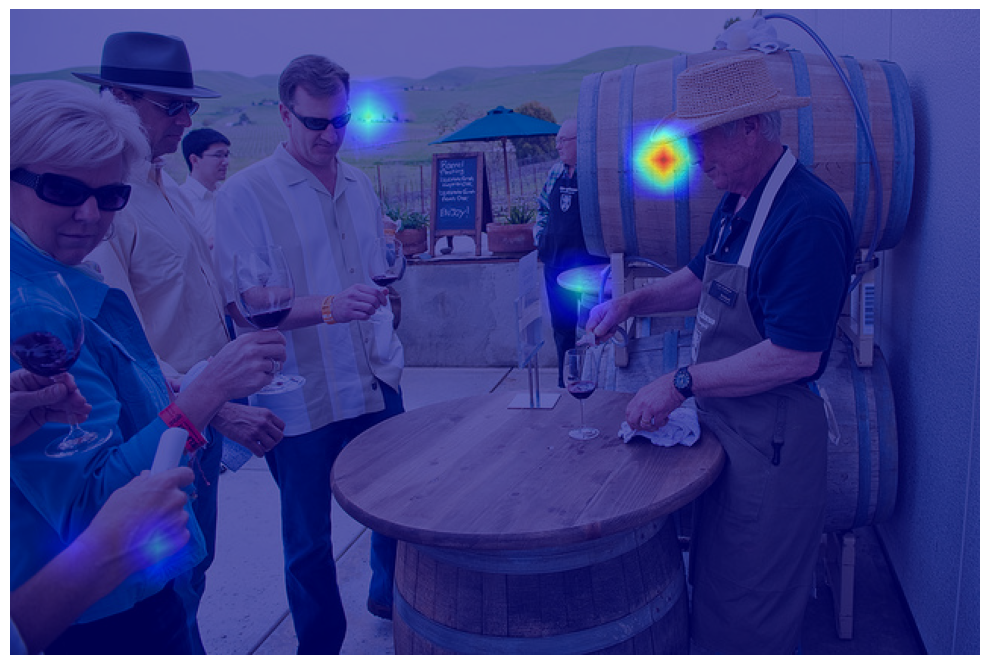}
    \caption{NTP}
  \end{subfigure}\hfill
  \begin{subfigure}[t]{0.24\textwidth}
    \vspace{0pt}
    \includegraphics[width=\textwidth]{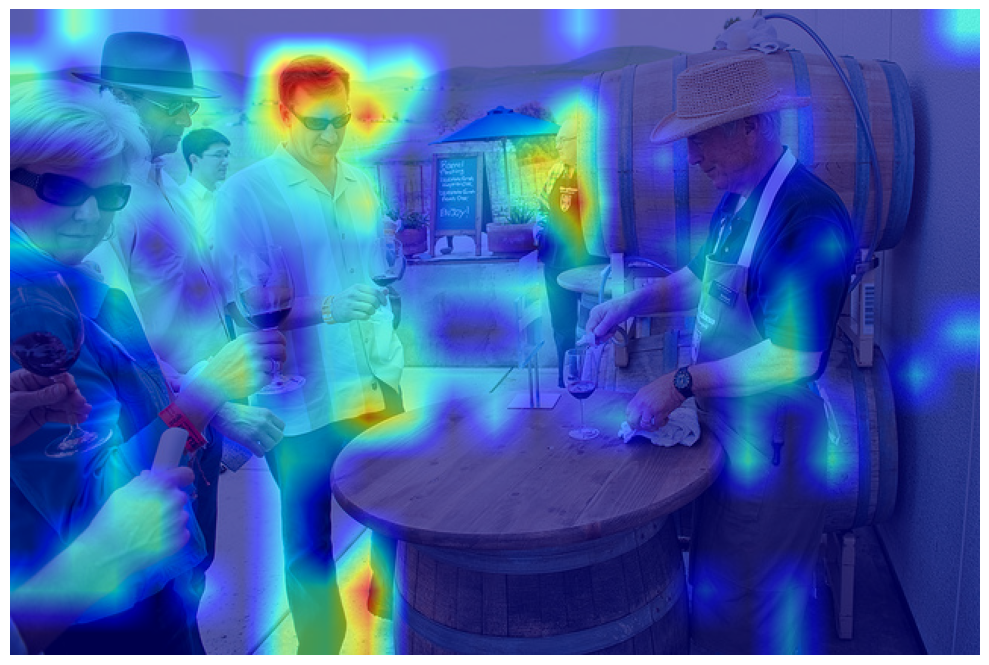}
    \caption{NTP+LLL}
  \end{subfigure}

  \caption{Example where object confidence maps do not perfectly localize the queried object. The target class is \textit{person}, with multiple instances occupying a large image area. From left to right: input image with target object category, base model, NTP-finetuned model, and NTP+LLL-finetuned model.}
  \label{fig:supp-confidence-limitation}
\end{figure*}

\Cref{fig:supp-confidence-limitation} shows a case where object confidence maps struggle to localize targets when multiple instances of the same category, such as \textit{person}, occupy much of the image. The base model exhibits broad activations, NTP finetuning improves focus, and NTP+LLL provides the most concentrated localization. Nevertheless, challenges remain in crowded scenes where objects are composed of a relatively large number of patches.

\captionsetup[subfigure]{labelformat=empty,font=footnotesize}

\bibliographystyle{splncs04}
\bibliography{references}